\documentclass[journal,twoside,web]{IEEEtran}
\usepackage{generic}
\usepackage{cite}
\usepackage{amsmath,amssymb,amsfonts}
\usepackage{algorithmic}
\usepackage{graphicx}
\usepackage{textcomp}
\usepackage{booktabs}
\usepackage{caption}
\usepackage{subcaption}
\usepackage{bbding}
\usepackage{algorithm}
\usepackage{makecell}
\usepackage[table]{xcolor}
\usepackage{algorithmic}\usepackage{multirow}
\def\BibTeX{{\rm B\kern-.05em{\sc i\kern-.025em b}\kern-.08em
    T\kern-.1667em\lower.7ex\hbox{E}\kern-.125emX}}
\begin{document}
\title{AdaTrendLike: Adaptive Hierarchical Detrending and Trend Likelihood Scoring for Time Series Anomaly Detection}

\author{Van Kwan Zhi Koh, Songnan Lin, Rongmo, Luo, Zhiping Lin, \IEEEmembership{Life Senior Member, IEEE}, and Bihan Wen, \IEEEmembership{Senior Member, IEEE}
\thanks{Van Kwan Zhi Koh, Songnan Lin, Zhiping Lin, and  Bihan Wen are with the School of Electrical and Electronic Engineering, Nanyang
Technological University, 50 Nanyang Ave, Singapore 639798. Email: vank0003@e.ntu.edu.sg, songnan.lin@ntu.edu.sg, EZPLin@ntu.edu.sg, bihan.wen@ntu.edu.sg}
\thanks{Rongmo, Luo is with Xylem Water Solutions Singapore Pte Ltd, Singapore 609935. Email: luo.rongmo@xylem.com}}
\maketitle

\begin{abstract}
The rapid proliferation of Industrial Internet of Things devices has led to increasingly large-scale and non-stationary time-series data, posing significant challenges for time-series anomaly detection (TSAD). Existing approaches often focus on reconstructing or forecasting observations while overlooking the role of trend dynamics, which constitute a major source of non-stationarity and contain valuable information for anomaly identification. To address this limitation, we propose AdaTrendLike, a TSAD framework that directly estimates trend likelihoods for anomaly detection. Specifically, we introduce Standardization Embedding to capture local statistical characteristics as latent representations and AdaDetrending to adaptively decompose input series into trend and detrended components. The detrended component is processed through a Fast Fourier Transform and a complex-valued convolutional neural network to learn normal periodic patterns, while TrendEst models the distribution of the trend component. Based on the estimated distribution, TrendLike computes anomaly scores by measuring the likelihood of observed trends, enabling effective detection of deviations from normal behavior. In addition to improving detection performance, the proposed framework provides feature-level interpretability by identifying anomalies through deviations in the estimated trend distribution. Extensive experiments on five public benchmark datasets demonstrate that AdaTrendLike consistently outperforms state-of-the-art TSAD methods, validating its effectiveness and robustness across diverse application scenarios.
\end{abstract}

\section{Introduction}
With the widespread deployment of Industrial Internet of Things devices, the volume of time series data has grown rapidly, creating a pressing need for robust Time Series Anomaly Detection (TSAD) tools for large-scale sensory data. In cybersecurity systems, malicious activity is flagged when the network exhibits unusual temporal behavior. In cyber-physical systems, which integrate digital and physical infrastructure, TSAD methods are utilized to detect physical anomalies caused by cyber intrusions in domains such as water treatment plants \cite{7469060} and unmanned aerial vehicles \cite{MoARNN-AM}. In IT infrastructures, monitoring systems track server metrics to detect anomalies indicative of system failures or resource bottlenecks \cite{abdulaal2021practical, su2019robust}. Despite differences in anomaly types across domains, the core challenge remains detecting deviations from normal behavior in temporally dependent, non-stationary time series.

Classical methods such as Local Outlier Factor (LOF), Isolation Forest (IF), and One-Class Support Vector Machine (OCSVM) \cite{liu2008isolation,tax2004support,breunig2000lof} are widely used. However, with the increasing scale and complexity of time series data, these methods struggle to capture temporal dependencies and high-dimensional feature interactions. Consequently, learning-based methods have emerged as a promising alternative by leveraging representation learning to model complex temporal and feature relationships \cite{song2024memto,xuanomaly,peng2022traffic}. These learning-based approaches typically rely on reconstruction errors to identify anomalies, which can be unreliable because normal variations or seasonal trends may be misclassified as anomalies. Moreover, this approach does not consider predictive uncertainties, which are crucial for informed and interpretable anomaly detection, such as identifying the features that contribute to anomalies.

Non-stationarity in real-world time series arises from multiple factors and can obscure the distinction between normal and abnormal patterns \cite{d3r}. One of these factors is the trend component, which represents the progressive long-term behavior of the data and can be leveraged in two main ways: as a dominant feature for modeling and as a means to detrend the series, promoting local stationarity. Trends can be extracted using traditional statistical models, but their effectiveness is limited because these models rely on the assumption of local stationarity \cite{hyndman2018forecasting}. In contrast, learning-based decomposition methods address this by incorporating mechanisms that extract the trend component within the inner blocks of the models \cite{d3r,fedformer,Wu2021AutoformerDT}. Although the inner blocks provide rich representations that capture underlying trends, these methods rely on trend separation mainly to improve reconstruction rather than explicitly exploiting the trend component for TSAD.

To address these challenges, we propose AdaTrendLike, an Adaptive Hierarchical Detrending and Trend Likelihood Scoring
method for TSAD. Our method first employs a Standardization Embedding (SE) module that encodes the local statistical properties of time series as latent embeddings. To further handle non-stationary behaviors, we propose Adaptive Detrending (AdaDetrending), a hierarchical module that progressively extracts trend components through multiple layers. At each layer, local trends are estimated using moving averages, and the residuals representing the detrended components are processed through a Fast Fourier Transform (FFT), followed by a complex-valued Convolutional Neural Network (CNN). This combination enhances the model’s capability to accurately capture normal behavior in time series exhibiting periodicity via frequency representations and local dependency via convolution. Finally, the trend components extracted from all layers are aggregated and modeled by the Trend Estimator (TrendEst) to estimate a global trend distribution, while the Trend Likelihood Scoring (TrendLike) module computes anomaly scores based on the likelihood of observed trends. By explicitly modeling the distribution of the trend component and performing anomaly detection through likelihood estimation, the proposed framework achieves improved robustness under non-stationary dynamics. Unlike existing methods that primarily exploit trend information to facilitate signal reconstruction, AdaTrendLike directly integrates trend modeling into the anomaly detection process. Specifically, deviations from the estimated trend distribution are treated as anomaly evidence, enabling effective detection of trend-related anomalies while enhancing robustness to non-stationary time-series data. The main contributions are as follows:
\begin{itemize}
    \item We propose AdaTrendLike, a framework comprising four modules: Standardization Embedding, AdaDetrending, TrendEst, and TrendLike. The framework hierarchically decomposes representations into trend and detrended components to address non-stationarity and estimates the trend distribution to capture deviations with feature-level interpretability.

    \item The AdaDetrending module performs hierarchical detrending using FFT, followed by a complex-valued CNN that effectively models normal temporal patterns, particularly in time series with periodic behavior.

    \item The TrendLike module computes anomaly scores for TSAD by leveraging the trend distribution estimated by TrendEst, enabling robust anomaly detection and interpretability through the quantification of deviations from the learned trend distribution.
\end{itemize}

The remainder of this paper is organized as follows. Section~\ref{sec:relatedwork} reviews the related work. Section~\ref{sec:meth} presents the proposed methodology. Section~\ref{sec:exp} presents the experimental results. Finally, Section~\ref{sec:conclude} concludes the paper.

\section{Related Work}
\label{sec:relatedwork}
We first review two common TSAD paradigms, namely deep one-class and reconstruction-based methods, followed by decomposition methods for handling non-stationarity and frequency-domain representation learning for periodic patterns, both of which are closely related to our method.

\subsection{Deep One-Class Methods}
Classical methods such as IF \cite{liu2008isolation}, OCSVM \cite{tax2004support}, and LOF \cite{breunig2000lof} have been widely used for unsupervised anomaly detection. Building on these foundations, advanced models like Deep-SVDD \cite{ruff2018deep}, which jointly train a neural network with Support Vector Data Description (SVDD) \cite{tax2004support} to enclose normal data in a minimal hypersphere, and THOC \cite{thoc}, which employs dilated recurrent neural networks with skip connections for multi-resolution features, have emerged. COUTA \cite{couta} further introduces uncertainty quantification and anomaly perturbation to enhance confidence in detection. However, detecting anomalies as deviations from the learned data manifold remains challenging in high-dimensional spaces due to the curse of dimensionality.
\subsection{Reconstruction-based Methods}
Deep learning methods often incorporate inductive biases, such as specific training objectives or architectural constraints, that favor the reconstruction of normal patterns. The resulting reconstruction errors are then used to identify anomalous segments. For instance, TranAD \cite{tranad} employs an adversarial training procedure to amplify reconstruction errors, while EvoAAE \cite{EvoAAE} optimizes a variational autoencoder–adversarial framework through particle swarm optimization. However, in both cases, the reliance on reconstruction-based objectives can lead to over-generalization, enabling anomalous patterns to be reconstructed too accurately and thereby masking their detection \cite{song2024memto}. MoARNN-AM \cite{MoARNN-AM} similarly uses a genetic algorithm to optimize the hyperparameters of recurrent neural network variants and attention mechanisms. However, it is designed for multi-class anomaly detection and relies on labeled anomalies, which are typically scarce in real-world settings \cite{nspr,couta}. Rather than depending on extensive hyperparameter optimization or labeled data, other methods introduce model-specific inductive biases. AT \cite{xuanomaly} amplifies normal–abnormal scores using the discrepancy between an association vector from a multi-head self-attention mechanism and a prior association vector based on a Gaussian distribution. MEMTO \cite{song2024memto} employs a Transformer encoder to extract latent representations, along with a gated memory module that is incrementally updated to capture prototypical normal patterns. Although these models are designed to reconstruct normal patterns effectively, they are trained exclusively on normal data. Consequently, variations arising from aleatoric uncertainty may be mistakenly flagged as anomalies. 

Instead of focusing on model design, some approaches improve detection performance by refining anomaly scoring criteria. DAGMM \cite{zong2018deep} employs a Gaussian Mixture Model for anomaly scoring, NSPR \cite{nspr} scores anomalies based on point-wise and sequential reconstruction deviations, and TranAD \cite{tranad} uses adversarial reconstruction errors. Although these strategies achieve strong empirical performance, they lack an explicit mechanism to account for deviations from the normal data distribution, limiting their robustness to non-anomalous variations.
\subsection{Decomposition of Dominant Components}
Learning from dominant decomposed components in time series analysis involves separating signals into key components, such as trend, seasonality, and residuals, to improve representation learning \cite{Wu2021AutoformerDT,fedformer,TFAD,d3r}. TransDe \cite{ZHANG2025107399}, SDDformer \cite{FU2024111671}, and Time-Frequency analysis based TSAD (TFAD) model \cite{TFAD} are methods that incorporate time-domain decomposition for TSAD. These methods primarily employ time-domain decomposition to separate temporal components, making the learned representations dependent on predefined decomposition assumptions, limiting their ability to capture complex temporal dynamics. Therefore, recent methods perform decomposition in the latent space. Autoformer \cite{Wu2021AutoformerDT} extracts trend components directly from latent representations, while D3R \cite{d3r} decomposes latent representations into trend and stable components before recombination. Although these models effectively learn trends to facilitate representation learning, they do not explicitly exploit trend information as an anomaly cue for TSAD. As a result, they fail to leverage the valuable information encoded in the trend component, where deviations from the underlying temporal trajectory can serve as strong indicators of anomalous behavior.

\begin{figure*}[!h]
    \centering \includegraphics[width=0.95\linewidth]{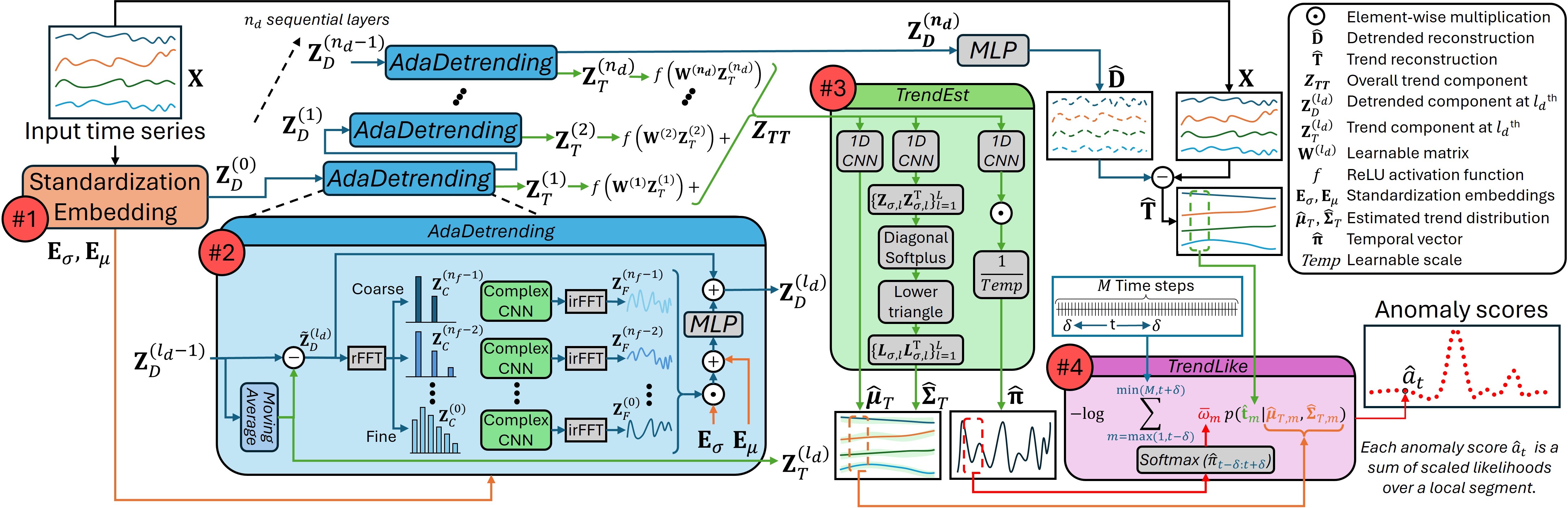} 
    \caption{
Overview of the AdaTrendLike framework. Red circles (\#1--\#4) indicate the sequential stages of the process. (\#1) Standardization Embedding: Maps input $\mathbf{X}$ to initial input representation $\mathbf{Z}_D^{(0)}$ and extracts standardized embeddings $\mathbf{E}_{\sigma}$, $\mathbf{E}_{\mu}$. (\#2) AdaDetrending: Removes the trend via moving average $\mathbf{Z}_T^{(l_d)}$ and captures periodic patterns via FreqCNN, yielding the detrended component $\mathbf{Z}_D^{(l_d)}$. (\#3) TrendEst: Aggregates layer-wise trends $\mathbf{Z}_{TT}$ to estimate distribution parameters ($\boldsymbol{\hat{\mu}}$, $\boldsymbol{\hat{\Sigma}}$) and the temporal vector $\boldsymbol{\hat{\pi}}$. (\#4) TrendLike computes the step-wise anomaly score \(\hat{a}_t\) from the weighted log-likelihood of the predicted trend \(\hat{\mathbf{t}}_m\) under the estimated distribution.
}
\label{overview}
\end{figure*}
\subsection{Frequency-domain Representation Learning}

Typically, FFT is applied directly to the input time series to transform time-domain signals into the frequency domain, enabling the capture of frequency-based patterns \cite{TFAD,ren2019time,li2025frequency,MAGE}. MAGE \cite{MAGE} transforms the input time series into the frequency domain using FFT, followed by a graph neural network to learn the feature topology. Similarly, TFAD method applies FFT to the decomposed components before modeling. Instead of raw or decomposed features, recent studies perform FFT on latent representations \cite{yi2024frequency,fedformer}. Fedformer \cite{fedformer} combines FFT with transformer self-attention, selecting random frequency components to capture dominant patterns. FreTS \cite{yi2024frequency} applies FFT to latent features and processes them with a complex-valued multilayer perceptron (MLP) to model global spectral components. Furthermore, existing approaches are primarily designed for TSF, with limited exploration in TSAD. In contrast, our method introduces frequency-learning biases to improve anomaly separability and employs a Complex CNN to capture frequency-domain periodicity and local temporal dependencies.

\section{Methodology}
\label{sec:meth}

\label{method}
\subsection{Problem Formulation}
Given a multivariate time series with a total length \(M\), we segment it into a set of sub-series using a rolling window of length \(L\) with a stride value \(s\), where each sub-series serves as the input and label for the model. Let \(\mathbf{X}=[\mathbf{x}_1, \mathbf{x}_2,\ldots,\mathbf{x}_L | \mathbf{x}_t \in \mathbb{R}^{d_x}]\) denote a subseries of the multivariate time series, where \(L\) is the length of the window and \(d_x\) is the number of multivariate features. This corresponds to a sequence of labels denoted as \(\mathbf{Y}=[y_1,\ldots,y_L | y_t\in{\mathbb{Z}_2}]\), where  \(y_t=1\) indicates that the time step is abnormal, and \(y_t=0\) indicates that the time step is normal. The model objective is to predict the abnormal time steps of \(\mathbf{X}\) by computing the anomaly scores for all time steps \(\hat{\mathbf{A}}=[\hat{a}_1, \ldots,\hat{a}_L|\hat{a}_t \in \mathbb{R}]\) and using a predefined threshold \(\zeta\), such that the predicted labels \(\bold{\hat{Y}} = [\hat{y}_1,\ldots,\hat{y}_L | \hat{y}_t \in \mathbb{Z}_2]\) are determined by \(\hat{y}_t=0\), if \(\hat{a}_t  < \zeta \) and \(\hat{y}_t=1\), if \(\hat{a}_t  \geq \zeta \).

\subsection{Overview of AdaTrendLike}
As shown in Fig. \ref{overview}, our method comprises four modules: Standardization embedding (SE), AdaDetrending, TrendEst, and TrendLike. The SE module first standardizes the input time series \(\mathbf{X}\) and projects the standardized data into a latent feature space \(\mathbf{Z}_D^{(0)}\). At the same time, this module encodes the statistical properties of the input into latent embeddings representing the mean \(\mathbf{E}_\mu\) and standard deviation \(\mathbf{E}_\sigma\). AdaDetrending is a hierarchical module that progressively extracts local trend components by using a moving average across multiple layers. At each layer, the detrended residuals are mapped to the frequency domain via a real FFT and processed using complex-valued CNNs, forming the FreqCNN module. After the final AdaDetrending layer, the resulting detrended component \(\mathbf{Z}^{(n_d)}_D\) is used to reconstruct the detrended reconstruction \(\hat{\mathbf{D}}\) via a Multilayer Perceptron (MLP). Meanwhile, the trend components \(\mathbf{Z}_T^{(l_d)}\) from each layer are aggregated using a learnable matrix \(\mathbf{W}\) through a nonlinear weighted sum to produce the overall trend component \(\mathbf{Z}_{TT}\). This component is passed to the TrendEst module, which estimates the parameters of the trend distribution, including the mean \(\boldsymbol{\hat{\mu}}_T\), covariance \(\boldsymbol{\hat{\Sigma}}_T\), and temporal vector \(\boldsymbol{\hat{\pi}}\). Finally, the TrendLike module utilizes these parameters to compute the anomaly scores \(\hat{\mathbf{A}}\), which are used to identify anomalous segments across time steps.

\subsection{Standardization Embedding}
\label{secsub:se}
Stationarity is essential for reliable time series analysis \cite{hyndman2018forecasting}. Although global standardization using training statistics is commonly applied, it may remove meaningful local patterns, such as magnitudes and variances, limiting model learning \cite{non-stationarytrans,patahmixerTSAD}. Inspired by \cite{non-stationarytrans}, we propose standardizing the input features based on the local input window rather than the entire training dataset and using the standardization parameters as embedding features. 

Given an input time series $\mathbf{X}$, we first compute the mean $\boldsymbol{\mu}_{in}$ and standard deviation $\boldsymbol{\sigma}_{in}$ along the feature dimension at each time step and use them to obtain the standardized input $\mathbf{\bar{X}}$. The standardized input $\mathbf{\bar{X}}$ is then projected by an MLP into the latent representation $\mathbf{Z}_D^{(0)} \in \mathbb{R}^{L \times d_m}$, where $d_m$ denotes the latent dimension. Unless otherwise specified, all MLPs in this paper consist of two linear layers with a ReLU activation function in between. To preserve local statistical information lost during standardization, $\boldsymbol{\mu}_{in}$ and $\boldsymbol{\sigma}_{in}$ are separately encoded using two dedicated MLPs, producing $\mathbf{E}_\mu \in \mathbb{R}^{d_m}$ and $\mathbf{E}_\sigma \in \mathbb{R}^{d_m}$, respectively. A Softplus activation is applied to $\mathbf{E}_\sigma$ to ensure positivity, enabling it to serve as a valid scaling factor. Overall, we denote this module as $SE(\mathbf{X})$, which outputs $\mathbf{Z}_D^{(0)}$, $\mathbf{E}_\mu$, and $\mathbf{E}_\sigma$. The embeddings $\mathbf{E}_\mu$ and $\mathbf{E}_\sigma$ are subsequently used in the AdaDetrending module to denormalize the latent representations at each layer.

\subsection{Adaptive Detrending}
Trends represent the long-term behavior of a time series, providing a reference for normal patterns and enabling detrending to mitigate non-stationarity. Unlike conventional approaches that extract trends directly in the time domain, the proposed method performs trend extraction in the projected latent space \cite{Wu2021AutoformerDT}, yielding a more expressive representation. Furthermore, our proposed hierarchical decomposition progressively learns multi-level trend representations and their corresponding detrended components, enabling the model to capture evolving trends across multiple temporal scales. At each detrending layer \(l_d\), the trend component \(\mathbf{Z}^{(l_d)}_T\) is obtained by applying a moving average \(MA(\cdot,k_{avg})\) with a window size \(k_{avg}\) along the time step dimension. This trend is then subtracted from the input \(\mathbf{Z}^{(l_d-1)}_D\), with edge values padded by replicating adjacent time steps to preserve the sequence length. As described,
\begin{equation}
\label{detrending}
\tilde{\mathbf{Z}}^{(l_d)}_D=\mathbf{Z}^{(l_d-1)}_D-\mathbf{Z}^{(l_d)}_T,
\end{equation}
where \(\mathbf{Z}^{(l_d)}_T = MA(\mathbf{Z}^{(l_d-1)}_D,k_{avg}) \in \mathbb{R}^{L \times d_m}\) is the trend component, \(\tilde{\mathbf{Z}}_D^{(l_d)} \in \mathbb{R}^{L \times d_m}\) is the detrended component, and the layer index \(l_d\) ranges from 1 to \(n_d\). Modeling time series in the frequency domain facilitates more effective identification of periodic patterns that may be difficult to detect in the time domain \cite{yi2024frequency, film,fedformer}. Motivated by the findings in \cite{yi2024frequency}, we employ a complex-valued CNN to process the frequency-domain representations at progressively increasing resolutions, ranging from coarse to fine. The CNN facilitates the learning of local structures inherent in the frequency domain, thereby enabling the model to emphasize dominant periodic components. This hierarchical processing enhances the characterization of normal behavioral patterns in time series exhibiting periodicity. The operation is represented as \( FreqCNN(\tilde{\mathbf{Z}}^{(l_d)}_D) \) with a residual connection. For each layer \(l_d\), the operation is defined as: 
\begin{equation}
\mathbf{Z}^{(l_d)}_D = FreqCNN(\tilde{\mathbf{Z}}^{(l_d)}_D) + \tilde{\mathbf{Z}}^{(l_d)}_D.
\end{equation}

We now formally define the FreqCNN module. Let $\tilde{\mathbf{Z}}^{(l_d)}_D \in \mathbb{R}^{L \times d_m}$ denote the detrended latent representation at the $l_d$-th layer. We first transform the input time series into the frequency domain using a real-valued fast Fourier transform (rFFT), which computes only the positive-frequency components of the real-valued input, denoted as $\mathcal{F}_{d_{l_f}}$ with frequency resolution $d_{l_f}$. This transformation is given by \(\mathbf{Z}_C^{(l_f)} = \mathcal{F}_{d_{l_f}}(\tilde{\mathbf{Z}}^{(l_d)}_D)\), 
where \( d_{l_f} = \frac{L}{2^{l_f}} \) denotes the frequency resolution at layer \( l_f \), for \( 0 \leq l_f < n_f \), with \( n_f \) being the total number of frequency resolution layers, and \(\mathbf{Z}^{(l_f)}_C\in \mathbb{C}^{(\frac{d_{l_f}}{2} +1) \times d_m}\) is the frequency-domain representation. After that, we apply a complex-valued CNN and inverse rFFT $\mathcal{F}^{-1}_{d_{l_f}}$, as described:
\begin{equation}
\mathbf{Z}_F^{(l_f)} =\mathcal{F}_{d_{l_f}}^{-1}\big(\text{ComplexCNN}(\mathbf{Z}_C^{(l_f)})\big),
\end{equation}
where $\text{ComplexCNN}(\cdot)$ denotes a complex-valued convolution defined under standard complex multiplication. The real and imaginary components are processed by two bias-free CNN branches with identical architecture, each followed by ReLU activation and configured with a grouping factor $g_{\text{cnn}}$ to reduce computational and memory overhead. The processed representation \(\mathbf{Z}^{(l_f)}_F \in \mathbb{R}^{d_{l_f} \times d_m }\) from each resolution is then concatenated along the \(d_{l_f}\) dimension, and standardization embeddings are applied to denormalize the latent representation, expressed as follows:
\begin{equation}
\label{denorm}
\mathbf{Z}_{\text{mul}}
=\text{Cat}_{d_{l_f}}\!\left(
\mathbf{Z}_F^{(0)}, \ldots, \mathbf{Z}_F^{(n_f-1)}
\right)
\odot
\left(\mathbf{1}\mathbf{E}_\sigma^\top \right)
+ \mathbf{1}\mathbf{E}_\mu^\top ,
\end{equation}
where $\text{Cat}_{d_{l_f}}(\cdot)$ denotes the concentration along the feature dimension $d_{l_f}$, $\mathbf{1} \in \mathbb{R}^{\sum_{i=0}^{n_f-1} \left\lfloor  \frac{L}{2^{i}} \right\rfloor}$, \(\odot\) denotes the element-wise product, and \(\mathbf{Z}_{\text{mul}}\in \mathbb{R}^{\sum_{i=0}^{n_f-1} \left\lfloor \frac{L}{2^{i}} \right\rfloor \times d_m} \). Finally, we apply an MLP to $\mathbf{Z}_{\text{mul}}$, mapping it back to the original latent shape and producing the output \(\mathbf{Z}_D^{(l_{f})}\in\mathbb{R}^{L \times d_m} \). Whereas the trend component \(\mathbf{Z}_T^{(l_d)}\), each layer captures temporal patterns at varying scales, progressing from local to global as the layer depth increases. 

With the trends extracted from each layer $\mathbf{Z}_T^{(l_d)}$, the overall trend representation $\mathbf{Z}_{TT}$ is obtained by integrating multi-scale trends via a nonlinear weighted summation: $ \mathbf{Z}_{TT} = \sum_{l_d=1}^{n_d} f\left(\mathbf{W}^{(l_d)} \mathbf{Z}_T^{(l_d)}\right)$, where \(\mathbf{W}^{(l_d)} \in\mathbb{R}^{L\times L}\) is a learnable matrix for the \(l_d\)-th layer and \(f(\cdot)\) denotes the ReLU function. After the AdaDetrending module, the detrended component $\mathbf{Z}^{(n_d)}_D$ is passed through an MLP to reconstruct the detrended signal $\hat{\mathbf{D}} \in \mathbb{R}^{L \times d_x}$, while the overall trend representation $\mathbf{Z}_{TT} \in \mathbb{R}^{L \times d_m}$ is fed into the TrendEst to estimate the parameters of the trend distribution.

\subsection{Trend Estimator}
TrendEst models the trend distribution as a multivariate Gaussian: $\mathbf{T} \sim \mathcal{N}(\boldsymbol{\hat{\mu}}_{T}, \boldsymbol{\hat{\Sigma}}_{T})$. Although this Gaussian assumption does not fully capture the true underlying distribution, it provides a flexible approximation for complex patterns while enabling robust uncertainty estimation. The overall trend $Z_{TT}$ is used to estimate the mean $\boldsymbol{\hat{\mu}}_{T}$ via a one-dimensional (1D) CNN, $\text{1DCNN}{\mu}$, formulated as follows:
\begin{equation}
\boldsymbol{\hat{\mu}}_{T}=\text{1DCNN}_{\mu}(\mathbf{Z}_{TT})\in \mathbb{R}^{L \times d_x},
\label{trendmu}    
\end{equation}
where $\text{1DCNN}_{\mu}$ denotes a two-layer, one-dimensional (1D) CNN applied along the temporal dimension with a kernel size of 3 and ReLU activation, whose output layer projects to a shape of $\mathbb{R}^{L \times d_x}$.

The estimation of $\boldsymbol{\hat{\Sigma}}_{T}$ requires positive definiteness and numerical stability, but incurs substantial memory and computational overhead. To address this, we leverage the overall trend $\mathbf{Z}_{TT}$ to predict a low-rank factor matrix $\mathbf{Z}_{\sigma}$. Then, a softplus transformation, $\operatorname{DiagonalSoftplus}(\cdot)$, is applied to the diagonal entries to ensure positivity. The lower-triangular component is then extracted using $\operatorname{tril}(\cdot)$, and the resulting matrix is used as a Cholesky factor to obtain a valid covariance matrix. As formulated,
\begin{equation}  
\label{eq:covariancecompute}
\mathbf{Z}_{\sigma}=\mathrm{Reshape}\big(\text{1DCNN}_{\sigma}(\mathbf{Z}_{TT})\big)\in \mathbb{R}^{L \times d_x \times r},
\end{equation}
\begin{equation}
\mathbf{L}_{\sigma ,l} = \mathrm{tril}\big(\operatorname{DiagonalSoftplus}(\mathbf{Z}_{\sigma,l} \mathbf{Z}_{\sigma,l}^\top)\big),
\end{equation}
\begin{equation}
\boldsymbol{\hat{\Sigma}}_{T,l} = \mathbf{L}_{\sigma ,l} \mathbf{L}_{\sigma ,l}^\top \in \mathbb{R}^{d_x \times d_x}, 
\quad l = 1, \dots, L,
\label{trendcovariance}
\end{equation}
where $\text{1DCNN}_{\sigma}$ shares the same architecture as $\text{1DCNN}_{\mu}$ but outputs a matrix in $\mathbb{R}^{L \times d_x r}$, which is reshaped into $\mathbb{R}^{L \times d_x \times r}$ to obtain a sequence of low-rank factor matrices $\mathbf{Z}_{\sigma,l} \in \mathbb{R}^{d_x \times r}$. We set $r = d_m/16 \ll d_m$, obtaining a compact low-rank parameterization that reduces memory and computation.

Since time series anomalies often persist or influence neighboring observations, step-wise anomaly scores should account for local temporal context. Therefore, we introduce a temporal vector $\boldsymbol{\hat{\pi}}$ to model the relative importance of neighboring time steps within a local window, formulated as:

\begin{equation}
\label{tempvector}
\hat{\boldsymbol{\pi}} 
= \frac{\text{1DCNN}_{\pi}(\mathbf{Z}_{TT})}{Temp} \in \mathbb{R}^L,
\end{equation}
where $\hat{\boldsymbol{\pi}}=\hat{\pi}_{1:L} = [\hat{\pi}_1, \hat{\pi}_2, \dots, \hat{\pi}_L]^\top$, $\text{1DCNN}_{\pi}$ denotes a two-layer, 1D CNN applied along the temporal dimension with a kernel size of 3 and ReLU activation, whose output layer maps to $\mathbb{R}^{L}$. Here, $\text{Temp} \in \mathbb{R}$ is a learnable scaling parameter initialized as $\sqrt{d_m}$, which controls the sharpness of the softmax distribution used in both the loss function and anomaly score computation. Together, $\boldsymbol{\hat{\mu}}_T$, $\boldsymbol{\hat{\Sigma}}_T$, and $\boldsymbol{\hat{\pi}}$ are used to compute the anomaly scores $\hat{\mathbf{A}}$ for each time step.
\subsection{Training Process}
The objective of the loss function is to jointly minimize the reconstruction loss of the detrended output, denoted as \(\mathcal{L}_R\), and maximize the likelihood of the estimated trend distribution, denoted as \(\mathcal{L}_{LL}\). The reconstruction loss is expressed as:
\begin{equation}
\label{lossrec}
\mathcal{L}_R = \frac{1}{L \cdot d_x} \, \Big\| \mathbf{X} - \mathbf{T} - \hat{\mathbf{D}} \Big\|_F^2,
\end{equation}
where $\mathbf{T} = MA(\mathbf{X}, k_{avg})$ denotes the ground truth trend. The \(MA(\cdot,k_{avg})\) is employed due to its simplicity and effectiveness in estimating underlying trends. The parameters of the multivariate Gaussian distribution are estimated by scaling the trend likelihoods with a time-dependent weight $\hat{w}_t$, expressed as:
\begin{equation}
\label{likeloss}
    \mathcal{L}_{LL}=\log\sum_{t=1}^{L}\hat{w}_t p(\mathbf{t}_{t}|\boldsymbol{\hat{\mu}}_{T,t},\boldsymbol{\hat{\Sigma}}_{T,t}),
\end{equation}
where $\mathbf{t}_t$ denotes the trend component at time step $t$, corresponding to the $t$-th row of $\mathbf{T}$, and $p(\mathbf{t}_t \mid \hat{\boldsymbol{\mu}}_{T,t} \hat{\boldsymbol{\Sigma}}_{T,t})$ denotes the probability density of a multivariate Gaussian distribution evaluated at \(\mathbf{t}_{t}\). The time-dependent weights are defined as
$\boldsymbol{\hat{w}} = \mathrm{softmax}\!\left( \hat{\pi}_{1:L} \right)$,
where the softmax is applied along the temporal dimension. The total loss function is defined as: \(\mathcal{L}_T=\mathcal{L}_R-\lambda \mathcal{L}_{LL}\), where \(\lambda>0\) is a hyperparameter that controls the trade-off between reconstruction loss and the accuracy of the trend distribution. The complete training procedure is outlined in  Algorithm \ref{training}.
\begin{algorithm}
\footnotesize
\begin{flushleft}
\caption{Training process}\label{training}
\textbf{Input:} $\mathbf{X} \in \mathbb{R}^{L \times d_x}$
\begin{algorithmic}
\STATE $\text{\textbf{Initialize:}}\quad\mathbf{Z}_{TT} = \textbf{0}$
\REPEAT
\STATE $\mathbf{Z}^{(0)}_D ,\mathbf{E}_{\mu}, \mathbf{E}_{\sigma} = \text{SE}(\mathbf{X})$
\FOR{$l_d = 1$ to $n_d$}
\vspace{0.2em}
\STATE $\mathbf{Z}_T^{(l_d)} = MA(\mathbf{Z}_D^{(l_d-1)},k_{avg})$ 
\vspace{0.2em}
\STATE $\tilde{\mathbf{Z}}_D^{(l_d)} = \mathbf{Z}_D^{(l_d-1)} - \mathbf{Z}_T^{(l_d)}$
\vspace{0.2em}
\STATE $\mathbf{Z}_D^{(l_d)} = FreqCNN(\tilde{\mathbf{Z}}_D^{(l_d)})+ \tilde{\mathbf{Z}}_D^{(l_d)}$
\STATE $\mathbf{Z}_{TT} = \mathbf{Z}_{TT} + f\left(\mathbf{W}^{(l_d)} \mathbf{Z}_T^{(l_d)}\right)$ 
\ENDFOR
\STATE $\hat{\mathbf{D}} = MLP(\mathbf{Z}_D^{(n_d)})$
\STATE $\hat{\boldsymbol{\mu}}_{T}, \hat{\boldsymbol{\Sigma}}_{T}, \hat{\boldsymbol{\pi}} = \text{TrendEst}(\mathbf{Z}_{TT})$
\UNTIL{$\mathcal{L}_T$ converges}
\end{algorithmic}
\end{flushleft}
\end{algorithm}

\subsection{TrendLike}
\label{sec:trendlike}
Unlike reconstruction-based anomaly scoring methods that rely on reconstruction errors to identify anomalies, we propose TrendLike, which leverages the estimated trend distribution. This approach improves sensitivity to subtle anomalies and better distinguishes between normal fluctuations and true abnormal behavior. The anomaly score $a_t$ at time step $t$ is defined as a weighted negative log-likelihood over a local temporal window:
\begin{equation}
\label{trendlikeeq}
\hat{a}_{t} = -\log \sum_{m=\max(1,t-\delta)}^{\min(M,t+\delta)} \bar{w}_{m}\, p\big(\hat{\mathbf{t}}_m \mid \boldsymbol{\hat{\mu}}_{T,m}, \hat{\boldsymbol{\Sigma}}_{T,m}\big),
\end{equation}
where $\hat{\mathbf{t}}_m$ denotes the $m$-th row of $\hat{\mathbf{T}} = \mathbf{X} - \hat{\mathbf{D}}$, corresponding to the trend reconstruction, $\delta$ denotes the window size, $t$ the time step, and $M$ the total length of the time series. The time-dependent weights $\boldsymbol{\bar{w}} = \mathrm{softmax}(\hat{\pi}_{t-\delta:t+\delta})$ define a normalized distribution over the local window $[t-\delta,\, t+\delta]$ along the time axis, thereby capturing the influence of neighboring time steps. This procedure is applied to all time steps up to $M$, rather than only within the training window length $L$. The resulting anomaly scores are denoted by $\hat{\mathbf{A}} = [\hat{a}_1, \hat{a}_2, \ldots, \hat{a}_M]$, from which anomalous time steps $\hat{\mathbf{Y}}$ are identified using a thresholding criterion.

\section{Experiments}
\label{sec:exp}
\subsection{Experiment Setup}
\subsubsection{Datasets}
This study utilizes five publicly available datasets from real-world applications: the Server Machine Dataset (SMD) \cite{su2019robust}, the Soil Moisture Active Passive (SMAP) and Mars Science Laboratory (MSL) datasets from NASA \cite{hundman2018detecting}, the Secure Water Treatment (SWAT) dataset \cite{7469060}, and the Pooled Server Metrics (PSM) dataset \cite{abdulaal2021practical}. Missing values are handled via forward filling, where the most recent non-null value is propagated to fill gaps. For SMD, MSL, and SMAP, which consist of multiple entities, the data are preprocessed by concatenating entities along the time-step dimension. Each dataset comes with predefined training and testing splits, with anomalies appearing exclusively in the test sets. Each dataset is provided with predefined training and testing partitions, with anomalous samples confined to the test partition. All datasets are normalized using min--max scaling to the range $[-1, 1]$, with scaling parameters estimated exclusively from the training data.

\subsubsection{Baseline Models}
We evaluate a diverse set of TSAD baselines spanning classical methods, including IF \cite{liu2008isolation}, OCSVM \cite{tax2004support}, and LOF \cite{breunig2000lof}; deep one-class approaches, including THOC \cite{thoc} and COUTA \cite{couta}; and reconstruction-based methods, including DAGMM \cite{zong2018deep}, TranAD \cite{tranad}, AT \cite{xuanomaly}, MEMTO \cite{song2024memto}, NSPR \cite{nspr}, and D3R \cite{d3r}. All models are evaluated on the provided datasets, except for D3R, which is designed to operate only on datasets with continuous-valued features. Experiments for THOC, AT, MEMTO, COUTA, NSPR, and D3R are conducted using publicly available implementations, while the remaining models are evaluated using the Multivariate TSAD toolkit \cite{liu2024mtad}. Each dataset has predefined training and testing splits, with details provided in Table~\ref{tab:datasetd}.

\begin{table}[!h]
\setlength{\tabcolsep}{2pt}
    \centering
        \caption{Details of the datasets}
    \begin{tabular}{c|ccccc} \toprule
        Datasets& SMD & MSL & SMAP & SWAT & PSM \\ \midrule
         Dimensions& 38 & 55 & 25 & 51 & 25 \\ 
         Entities& 28 & 27 & 55 & 1 & 1 \\
        Number of train steps  & 708405 & 58317 & 140825 & 495000 & 132481 \\ 
         Number of test steps & 708420 & 73729 & 444035 & 449919 & 87841 \\ 
        \text{Anomaly steps}/\text{Test steps}& 4.16\% & 10.48\% & 12.83\% & 12.14\% & 27.76\% \\ \bottomrule
    \end{tabular}

    \label{tab:datasetd}
\end{table}

\subsubsection{Evaluation Metrics}
Anomalous time steps are determined by a predefined threshold \(\zeta\). The predicted labels \(\bold{\hat{Y}} = [\hat{y}_1,\ldots,\hat{y}_L | \hat{y}_t\in{\mathbb{Z}_2}]\) are determined by \(\hat{y}_t=0\), if \(\hat{a}_{t}  < \zeta \) and \(\hat{y}_t=1\), and if \(\hat{a}_{t} \geq \zeta \), where \(\zeta\) is obtained by iteration via grid search to maximize the F1-score \cite{nspr,song2024memto,d3r}, referred to as the Best F1-score. A common evaluation practice, known as point adjustment, treats a predicted anomaly as correct if it falls within any truth anomalous segment \cite{song2024memto}. However, several studies question this practice, as it enables even trivial or adversarial algorithms to achieve inflated performance scores \cite{nspr,d3r}. Consequently, we evaluate our model strictly based on its raw predictions, without applying point adjustment. In addition, we report threshold-free performance using the Area Under the Receiver Operating Characteristic curve (AUC) to provide a complementary evaluation.

\subsubsection{Implementation Details and Hyperparameter Selection}
Experiments are conducted using the PyTorch library~\cite{paszke2019pytorch} on a server equipped with an RTX 3090 GPU. The model is trained for 100 epochs with a batch size of 64, employing the Adam optimizer with a learning rate of \(1 \times 10^{-4}\). Early stopping is applied when the validation loss does not decrease for two consecutive epochs. For all public datasets, we use the training portion as defined by the dataset providers, which is assumed to be free of anomalies. From this training set, we adopt a 90/10 split to create training and validation subsets. We set the input window length equal to the anomaly scoring window, \(L = \delta\), and a stride of \(s = L/4\) is applied across all datasets. The remaining hyperparameters, along with their search ranges and selected values, are summarized in Table~\ref{tab:hyperparam_full}. For a fair comparison with benchmark models, labeled anomalies are used for hyperparameter tuning, following prior work \cite{nspr, d3r, song2024memto}. A detailed sensitivity analysis is provided in Section~\ref{sec:hyper}, offering practical guidance on which hyperparameters are most critical for achieving strong performance.

\begin{table}[!h]
\centering
\setlength{\tabcolsep}{2.5pt}
\caption{Search space and selected hyperparameters}
\begin{tabular}{llccccc}
\toprule
\textbf{} &Search space & SMD &MSL& SMAP & SWAT&PSM \\
\midrule
$L$& $\{64,128,256,512,1028\}$ & 128 & 128 & 128 & 128 & 128 \\
$d_{m}$& $\{64,128,256,512,1028\}$ & 256 & 512 & 256 & 256 & 64 \\
$k_{avg}$ & $\{3,5,7,9,11\}$ & 3 & 3 & 3 & 3 & 3 \\
$k_f$ & $\{3,5,7,9,11\}$ & 3 & 3 & 3 & 3 & 3 \\
$n_f$ & $\{1,2,4,6,8\}$& 4& 4 & 4 & 4& 4 \\
$n_d$ & $\{1,2,4,8,16\}$ & 4 & 4 & 1 & 4& 4 \\
$g_{cnn}$ & $\{1,8,16,32,64\}$ & 16 & 16 & 64 & 16 & 16 \\
$\lambda$ & $\{1e^{-3},0.1,1,10,100\}$  & 1 & 1 & 1 & 1 & 1 \\
\bottomrule
\end{tabular}
\label{tab:hyperparam_full}
\end{table}
\subsection{Model Performance}
This section presents quantitative evaluations against baseline models, followed by visualization analyses that provide additional insights into the results. For visualization consistency, in Figs.~\ref{fig:limit}--\ref{fig:stationarity}, \ref{scoringcriteriaplots}, and \ref{fig:noisecompare}, black denotes the trend reconstruction $\hat{\mathbf{T}} = \mathbf{X} - \hat{\mathbf{D}}$, blue denotes the estimated trend mean $\hat{\boldsymbol{\mu}}_T$ with confidence intervals $\mathrm{diag}(\hat{\boldsymbol{\Sigma}})$, and red shading indicates ground-truth anomaly segments $\mathbf{Y}$.
\subsubsection{Main results} The main results are summarized in Table \ref{tab:mainexp}, where \textbf{bold} values indicate the best results. Our proposed AdaTrendLike achieves the best performance on 4 out of 5 datasets, demonstrating its effectiveness in TSAD. While deep learning approaches have been proposed, classical methods such as IF and LOF remain strong baselines. Deep One-Class methods, such as THOC and COUTA, which extend classical architectures, offer only marginal gains over classical methods. AT and MEMTO employ inductive biases to foster diverse representations; however, this may cause over-generalization that negatively impacts anomaly detection accuracy. DAGMM and NPSR employ unique anomaly scoring mechanisms that yield strong performance, highlighting the crucial role of effective anomaly scoring strategies. NPSR achieves highly competitive results by leveraging sequential pattern comparison to characterize normal behavior. Similarly, our model specifically utilizes trend information.  D3R performs decomposition but relies on reconstruction loss, whereas our method employs TrendLike for anomaly scoring, achieving better results.  

Notably, AdaTrendLike achieves the best performance on four out of five datasets, demonstrating strong effectiveness for TSAD. On SMD, the performance gain is less pronounced, likely due to its highly imbalanced anomaly distribution, with only 4.16\% anomalous time steps (Table~\ref{tab:datasetd}). As shown in Fig.~\ref{fig:limit}, many anomalous observations closely follow the underlying trends of normal data, resulting in weak anomaly evidence. In addition, SMD exhibits substantially shorter anomaly segments, with a mean length of $90$, compared to MSL ($216$), SMAP ($826$), SWaT ($1561$), and PSM ($339$). Since our model focuses on learning trends that capture long-term temporal patterns, short anomaly segments provide limited evidence for reliably estimating trend deviations, thereby reducing the effectiveness of TrendLike-based anomaly scoring. Nevertheless, AdaTrendLike still achieves the second-best AUC on SMD, indicating competitive ranking performance. Overall, these results highlight the advantages of the proposed hierarchical detrending framework, frequency-domain modeling, and TrendLike-based anomaly scoring.

\begin{table}[h]
        \setlength{\tabcolsep}{1pt}
        \captionof{table}{ROC and Best F1-Scores of all models}
        \begin{tabular}{c|cc|cc|cc|cc|cc}
            \toprule
             \multirow{2}{*}[-0.3em]{Models}&\multicolumn{2}{c}{SMD}&\multicolumn{2}{c}{MSL}&\multicolumn{2}{c}{SMAP}&\multicolumn{2}{c}{SWAT}&\multicolumn{2}{c}{PSM}\\
             \cmidrule(lr){2-11}
            &AUC&F1&AUC&F1&AUC&F1&AUC&F1&AUC&F1\\
            \midrule
            IF\cite{liu2008isolation}&0.664&0.176&0.603&0.252&0.646&0.312&0.845&0.736&0.685&0.500\\ 
            OCSVM\cite{tax2004support}&0.623&0.228&0.525&0.241&0.393&0.227&0.881&0.758&0.573&0.465\\
            LOF\cite{breunig2000lof}&0.645&0.124&0.556&0.216&0.623&0.283&0.888&0.774&0.728&0.583\\
            THOC\cite{thoc}&0.653&0.229&0.586&0.218&0.412&0.227&\textbf{0.893}&0.754&0.707&0.501\\ 
            COUTA\cite{couta}&0.660&0.199&0.557&0.212&0.526&0.259&0.751&0.502&0.693&0.529\\ 
            DAGMM\cite{zong2018deep}&0.689&0.180&0.600&0.229&0.513&0.270&0.862&0.749&0.660&0.484\\ 
            TranAD\cite{tranad}&0.655&0.171&0.549&0.192&0.572&0.255&0.888&0.819&0.733&0.569\\ 
            AT\cite{xuanomaly}&0.400&0.044&0.445&0.166&0.606&0.264&0.478&0.097&0.455&0.368\\ MEMTO\cite{song2024memto}&0.503&0.038&0.495&0.030&0.511&0.113&0.767&0.527&0.502&0.061\\
            D3R\cite{d3r}&0.714&0.201&-&-&-&-&0.343&0.218&0.590&0.443\\ 
            NSPR\cite{nspr}&\textbf{0.833}&\textbf{0.386}&0.653&0.283&0.534&0.248&0.890&0.847&0.792&0.645\\ \textbf{Ours}&0.722&0.203&\textbf{0.684}&\textbf{0.310}&\textbf{0.658}&\textbf{0.377}&0.884&\textbf{0.850}&\textbf{0.820}&\textbf{0.687}\\
            \bottomrule
        \end{tabular}
        \label{tab:mainexp}
\end{table}

\subsubsection{Visualization Analysis} 
\begin{figure}[h]
    \centering
    \begin{subfigure}[b]{0.425\linewidth}
        \centering
    \includegraphics[width=1\linewidth]{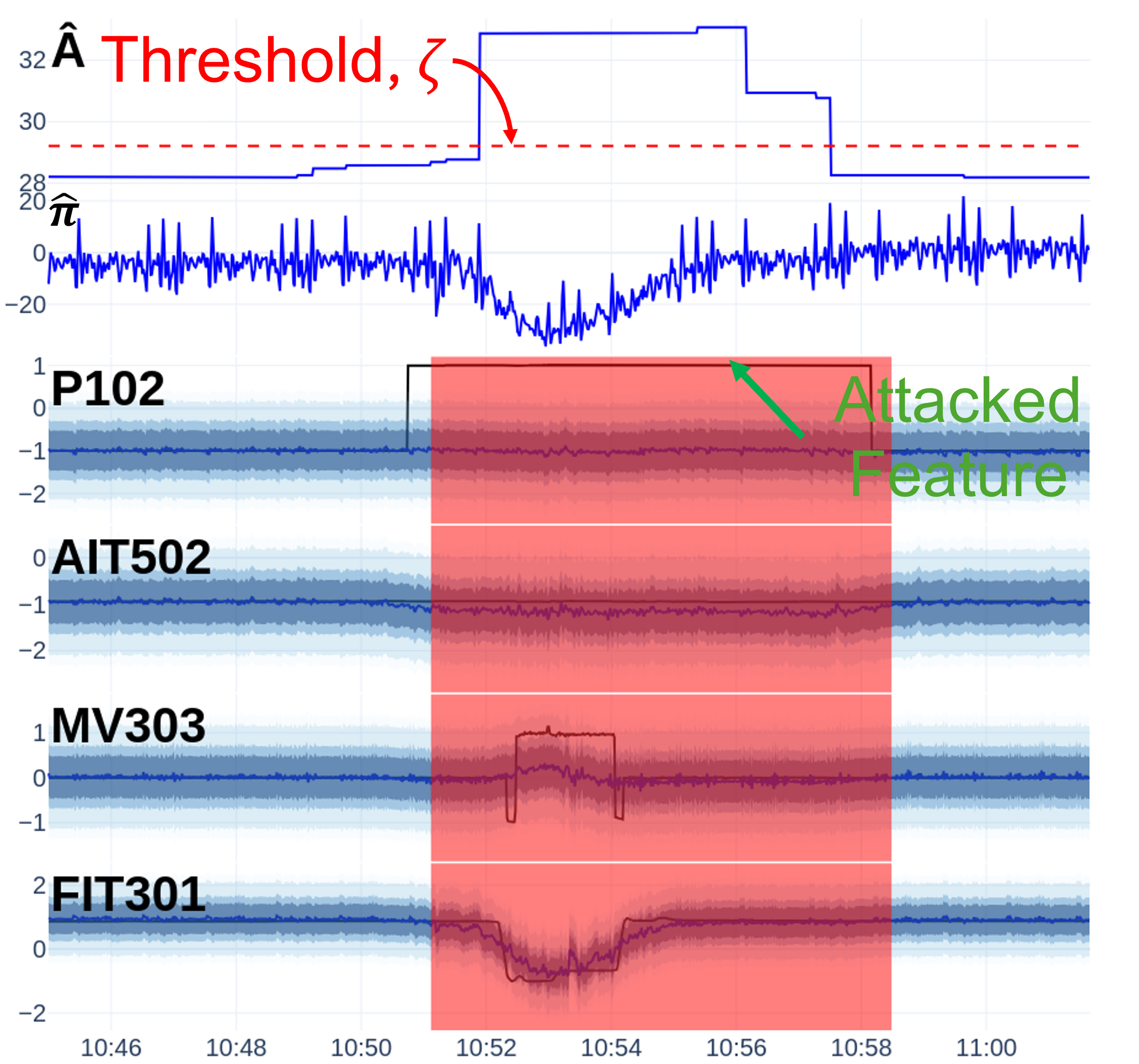}
    \caption{Scenario 1}
    \label{fig:case1}
    \end{subfigure}
    \begin{subfigure}[b]{0.425\linewidth}
        \centering
    \includegraphics[width=1\linewidth]{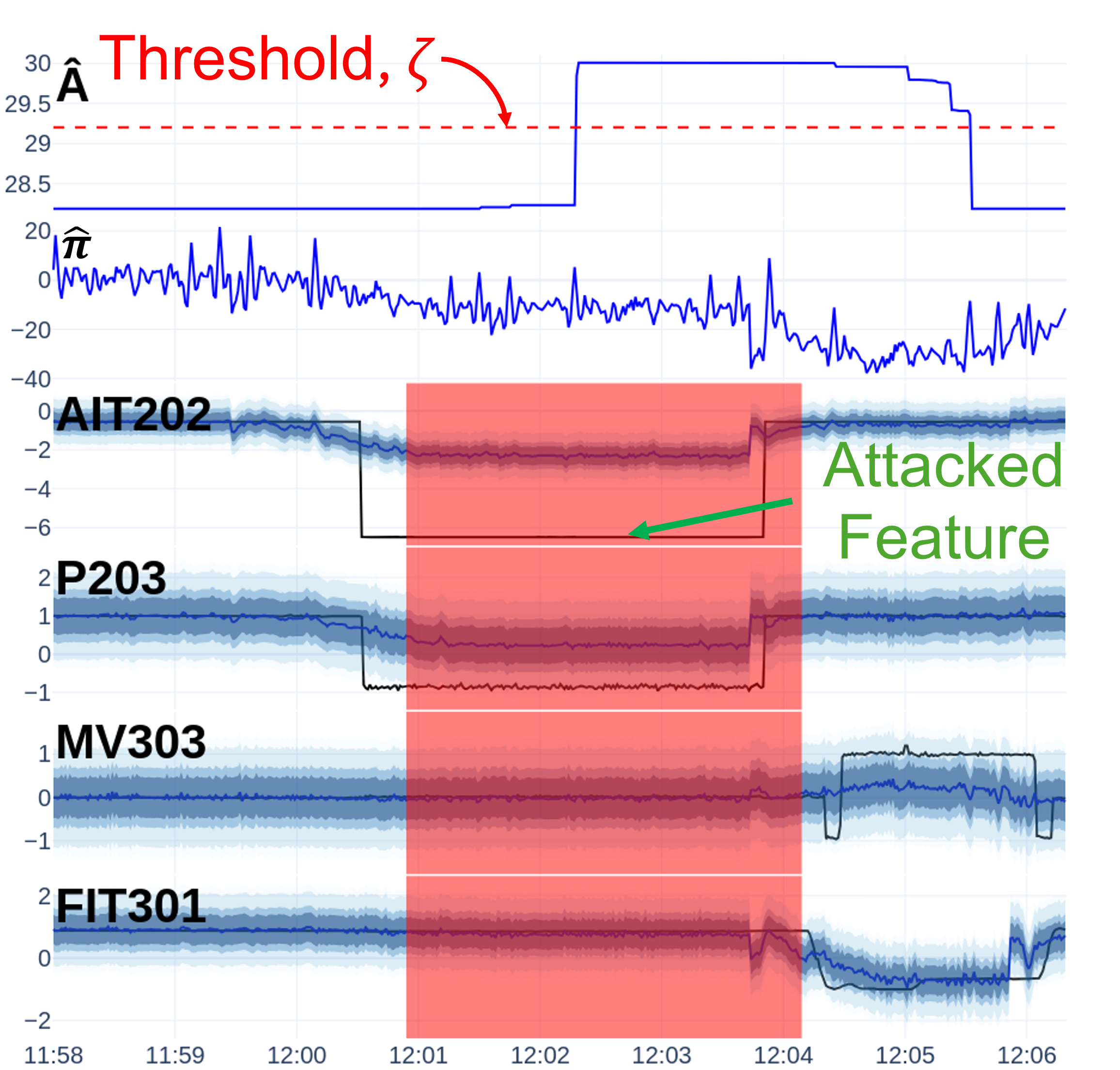}
    \caption{Scenario 2}
    \label{fig:case2}
    \end{subfigure}
    \begin{subfigure}[b]{0.425\linewidth}
        \centering
    \includegraphics[width=1\linewidth]{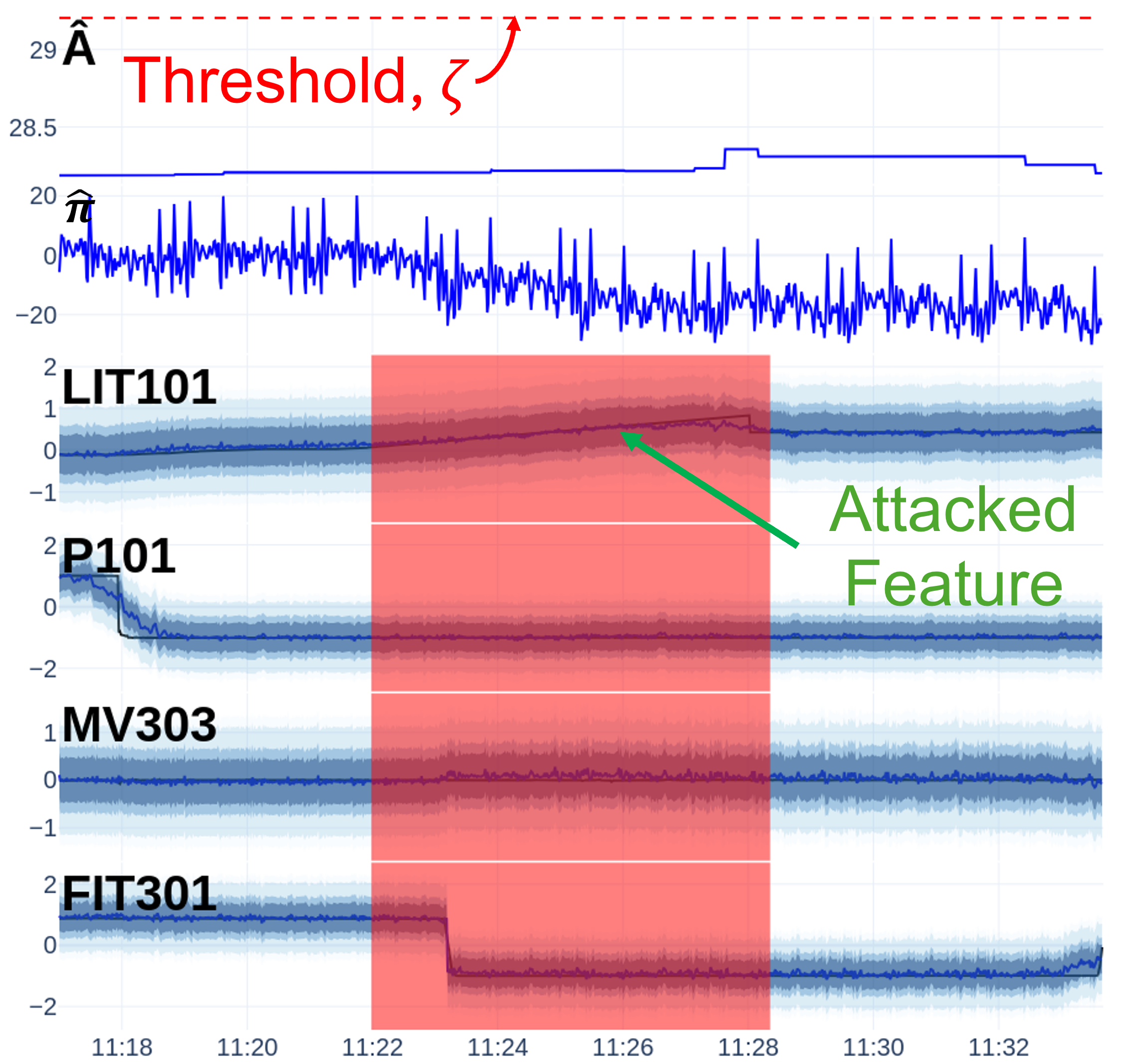}
    \caption{Scenario 3}
    \label{fig:case3}
    \end{subfigure}
    \begin{subfigure}[b]{0.425\linewidth}
        \centering
    \includegraphics[width=1\linewidth]{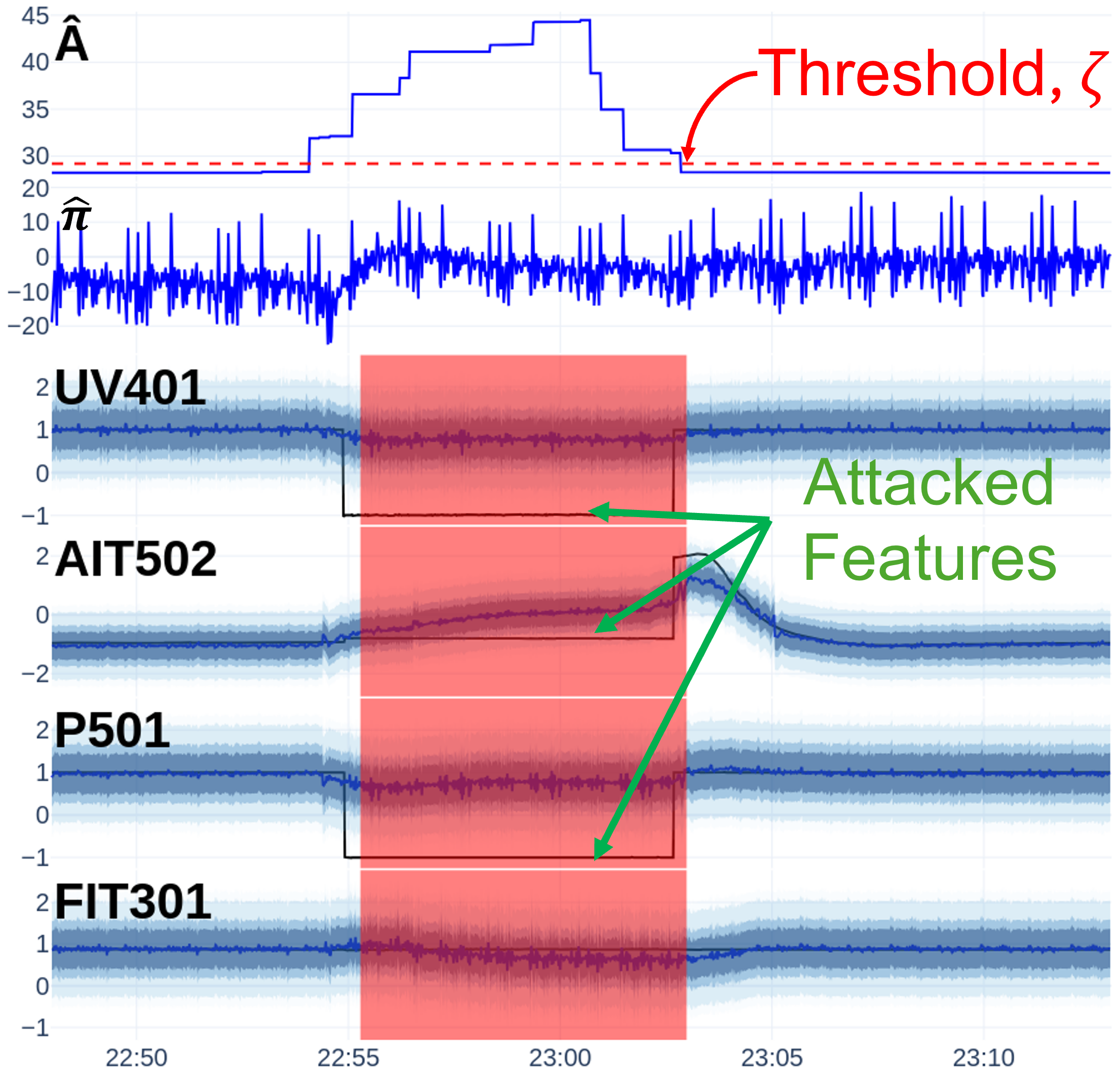}
    \caption{Scenario 4}
    \label{fig:case4}
    \end{subfigure}
    \caption{Visualization of anomaly scenarios for SWAT dataset}
    \label{fig:scenarios}
\end{figure}

Anomalies may arise from various features and time steps, but identifying which features are responsible is challenging due to the high dimensionality of the input data. Our model addresses this by providing a visual comparison between the estimated trends and trend reconstruction. From the SWAT dataset, four labeled scenarios, with feature names (e.g., P102) following the dataset documentation, are summarized in Table~\ref{tab:scenarios}. ‘Attacked features’ indicate features subjected to external changes, while ‘Expected feature changes’ indicate features for which changes are anticipated. As shown in Fig.~\ref{fig:scenarios}, scenarios 1, 2, and 4 show deviations between estimated trends and expected feature changes. Notably, MV303 in scenario 1 shows an unexpected deviation, suggesting a link to the attacked feature. For scenario 3, which involves gradual changes, our model does not detect this anomaly segment, likely because the trend-learning mechanism treats such changes as normal behavior.

\begin{figure}[h!]
    \centering
    \begin{subfigure}[b]{0.425\linewidth}
        \centering
    \includegraphics[width=1\linewidth]{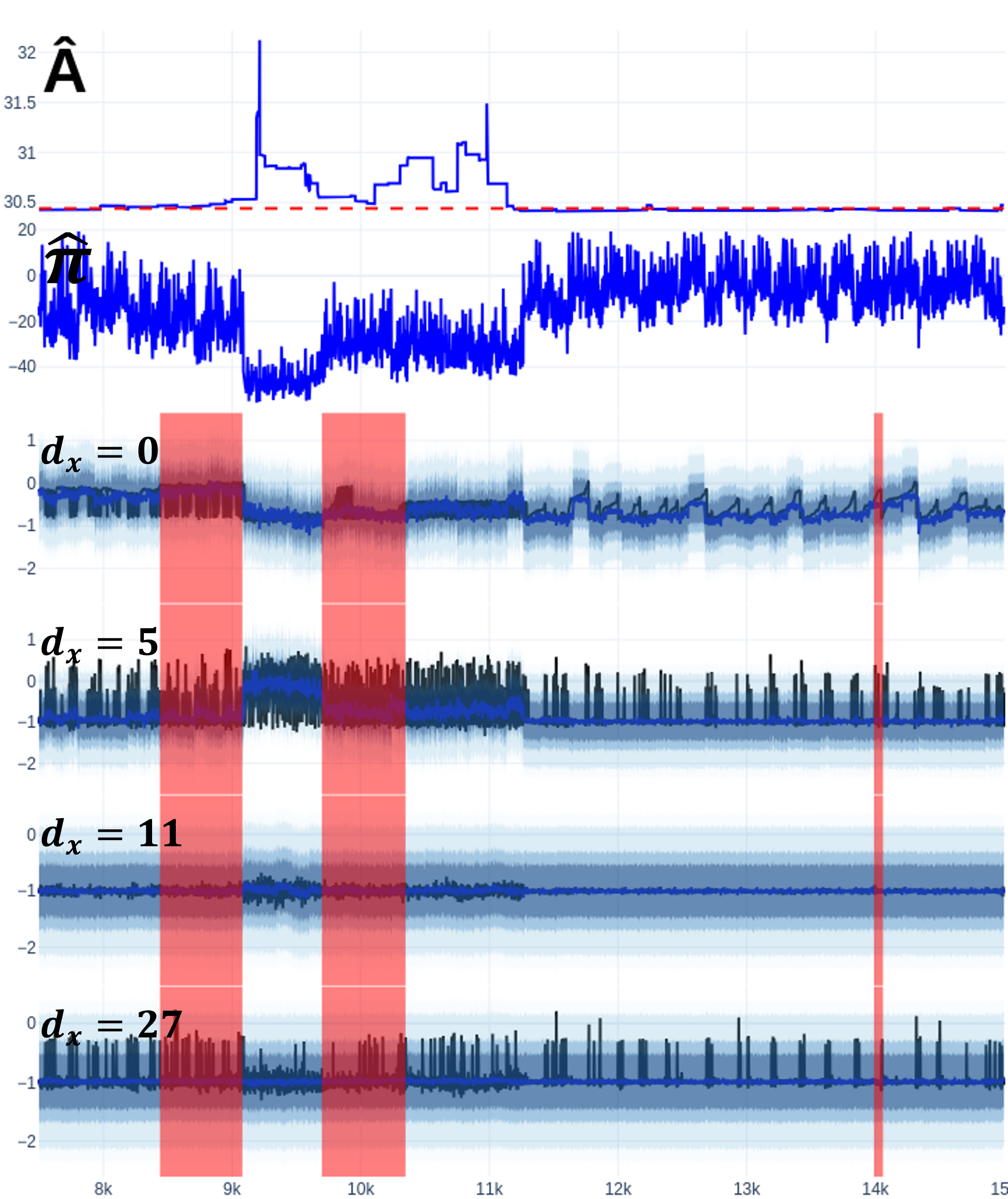}
    \caption{MSL }
    \label{fig:mslest}
    \end{subfigure}
    \begin{subfigure}[b]{0.425\linewidth}
        \centering
    \includegraphics[width=1\linewidth]{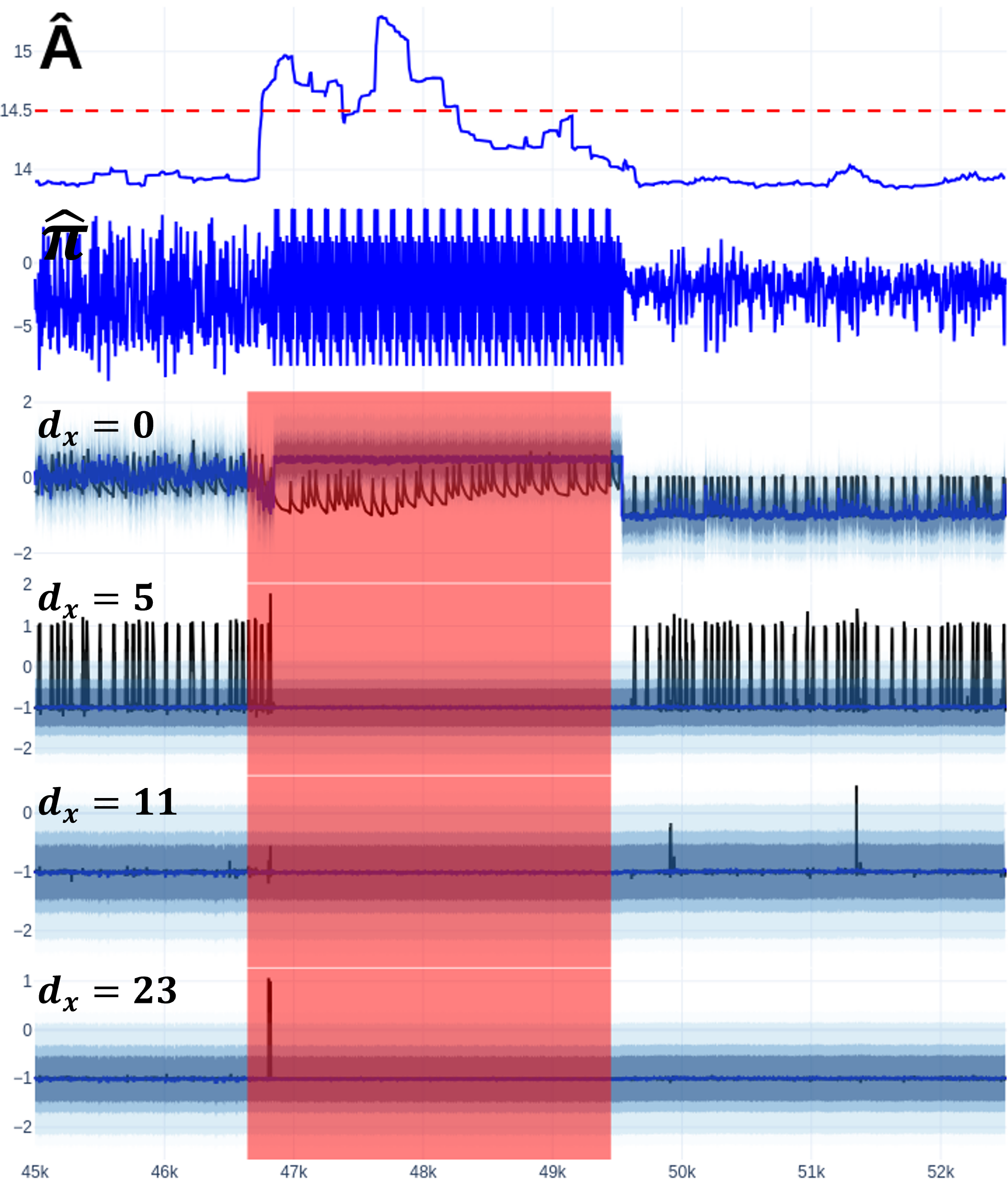}
    \caption{SMAP}
    \label{fig:smapest}
    \end{subfigure}
    \begin{subfigure}[b]{0.425\linewidth}
        \centering
    \includegraphics[width=1\linewidth]{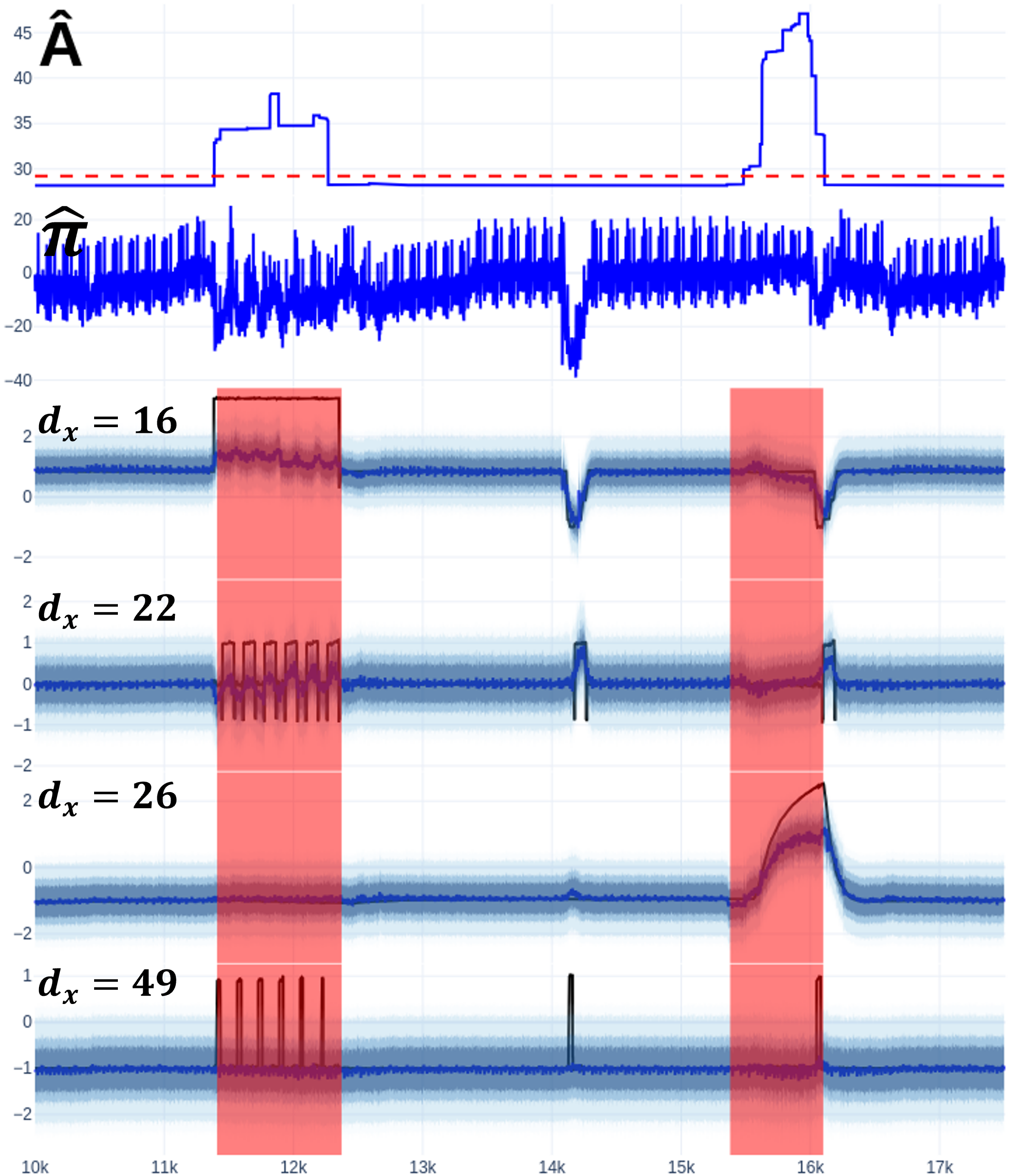}
    \caption{SWAT}
    \label{fig:swatest}
    \end{subfigure}
    \begin{subfigure}[b]{0.425\linewidth}
        \centering
    \includegraphics[width=1\linewidth]{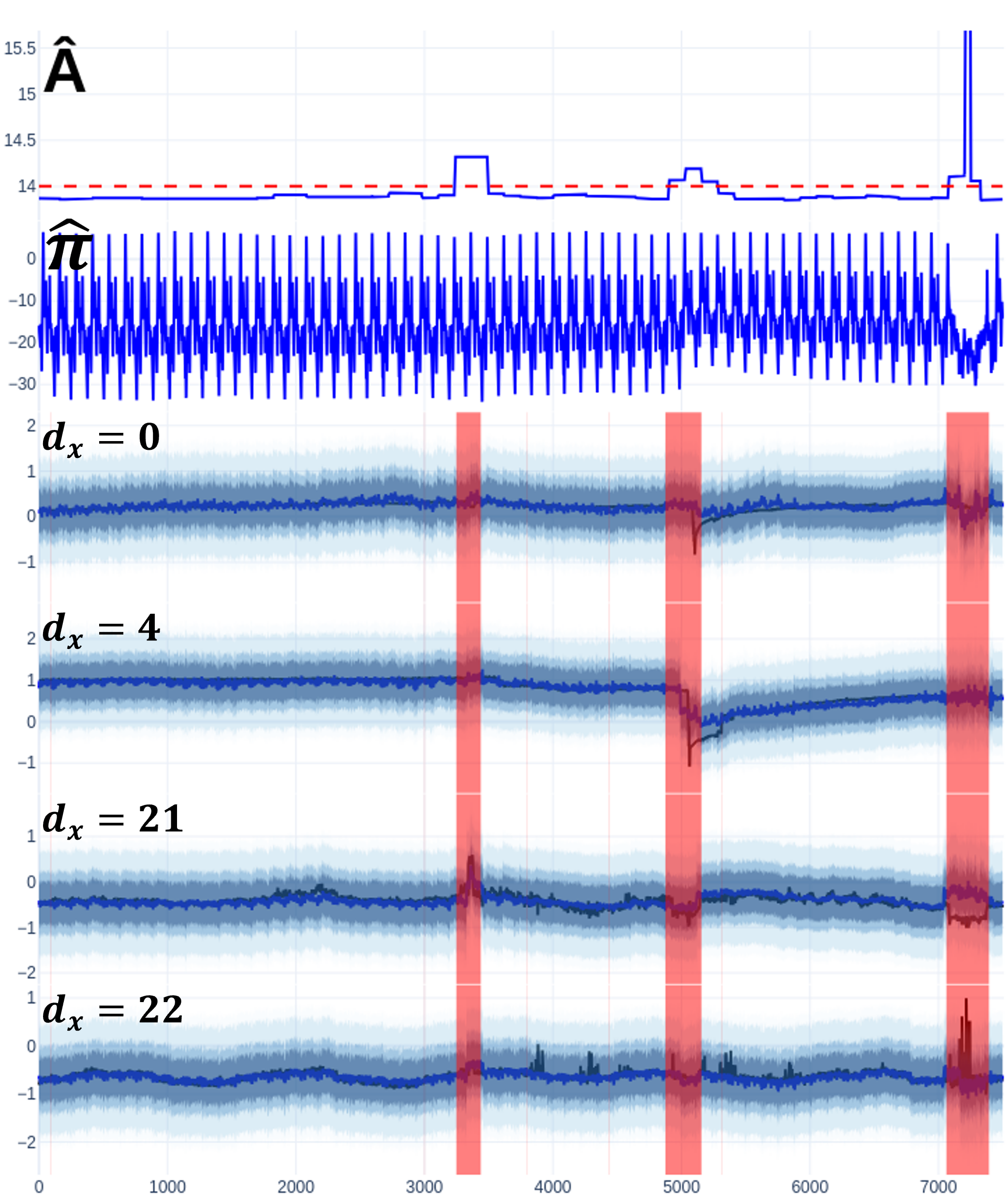}
    \caption{PSM}
    \label{fig:psmest}
    \end{subfigure}
    \caption{Visualization of trend distribution}
    \label{fig:predictions}
\end{figure}

\begin{figure}[!h]
    \centering   \includegraphics[width=0.85\linewidth]{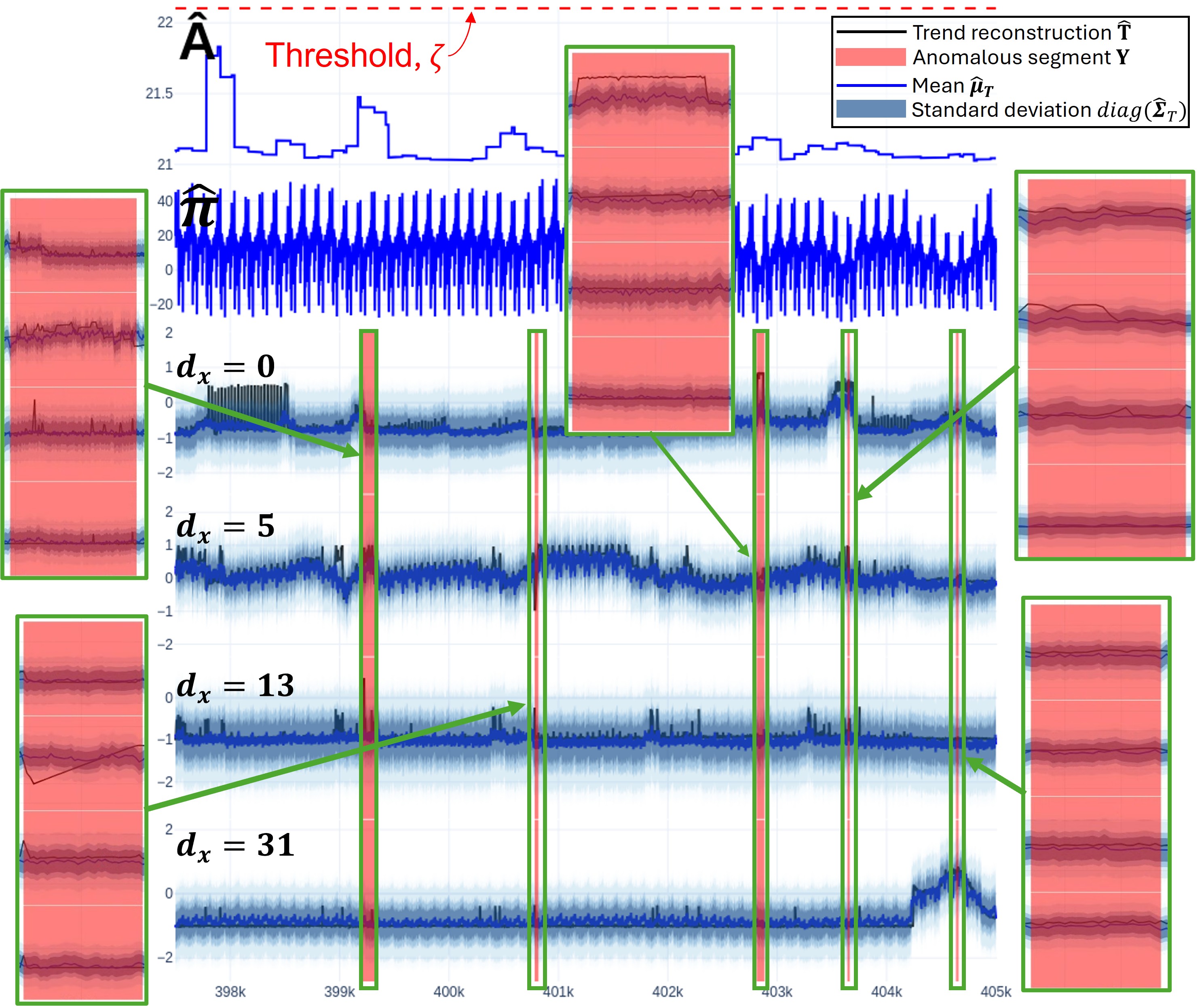}
    \caption{Trend distribution for selected SMD features}
    \label{fig:limit}
\end{figure}

Based on these scenarios, practitioners can assess anomalies from deviations along the estimated trend, highlighting contributing features and detection likelihood. For further analysis, trend estimation on other datasets are presented in Fig.~\ref{fig:predictions}. The MSL and SMAP datasets contain frequent short-term spikes, whereas the SWAT and PSM datasets exhibit smoother patterns. Our model captures these spikes within the estimated trend distribution, adapting to such patterns. The temporal vector provides a global representation of the features within a segment, capturing patterns such as sudden drops, mean shifts, and high-frequency fluctuations. During anomalous segments, it highlights noticeable changes in these characteristics, which can serve as indicators of anomaly occurrence.

\begin{table}[h]
\setlength{\tabcolsep}{1.5pt}
\footnotesize
    \caption{Labeled anomaly scenarios for SWAT dataset}
    \centering
    \begin{tabular}{c|c|c}
    \toprule
         Scenarios&Attacked features & Expected feature changes \\
    \midrule
         1&P102 & P102 is expected to switch from off to on. \\
         2&AIT202 & AIT202 and P203 are expected changes. \\
         3&LIT101 & LIT101 exhibits gradual changes. \\
         4&UV401,AIT502,P501 & All listed features are expected to change. \\
    \bottomrule
    \end{tabular}
    \label{tab:scenarios}
\end{table}

\subsection{Ablation Study}
This section presents the ablation study results using the best F1-scores, unless otherwise specified.

\subsubsection{Module Analysis}
\begin{figure}[!h]
    \centering
    \begin{subfigure}[b]{0.425\linewidth}
        \centering
        \includegraphics[width=\linewidth]{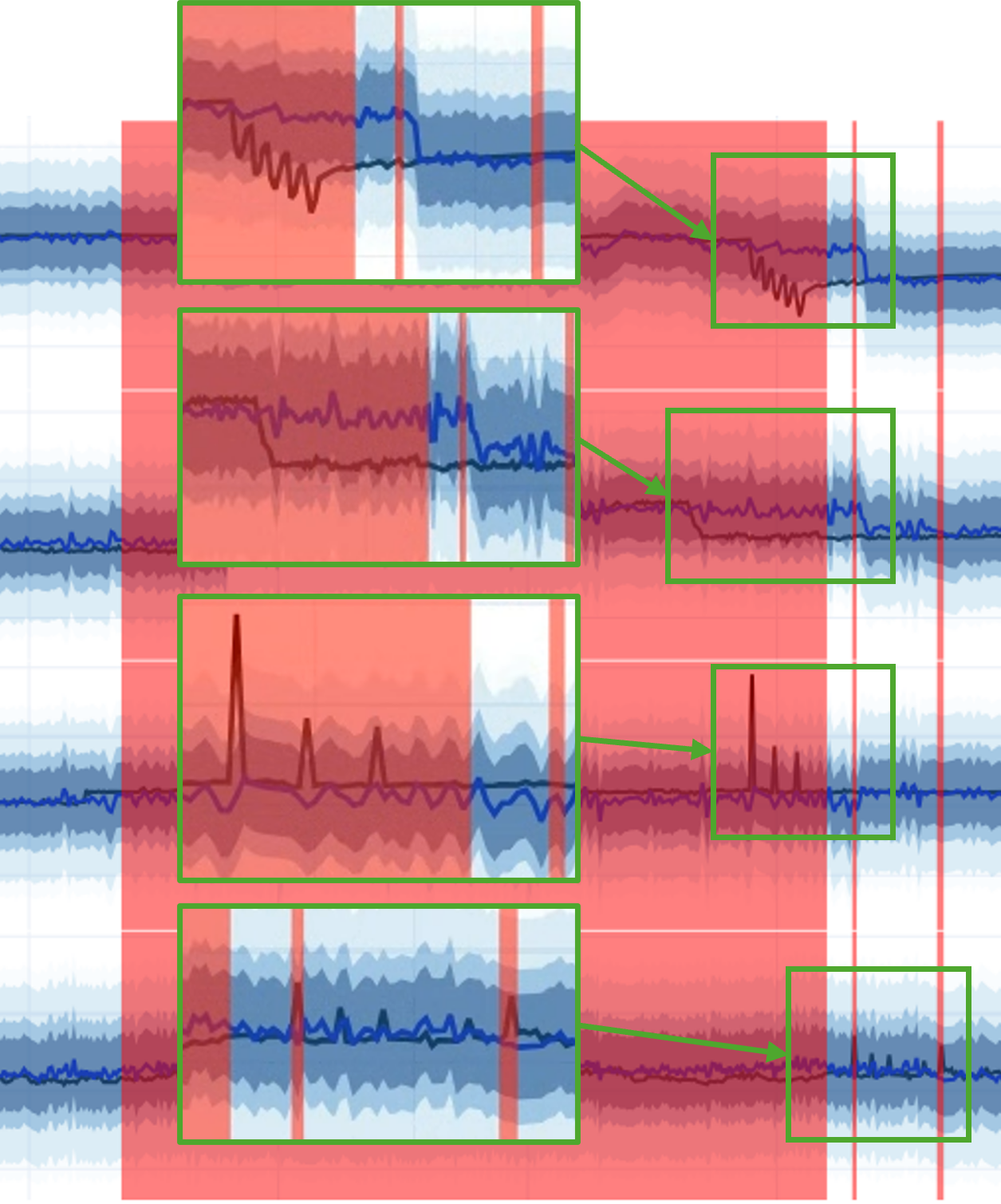}
        \caption{Ours}
        \label{fig:ours_exdetrend}
    \end{subfigure}
    \begin{subfigure}[b]{0.425\linewidth}
        \centering
        \includegraphics[width=\linewidth]{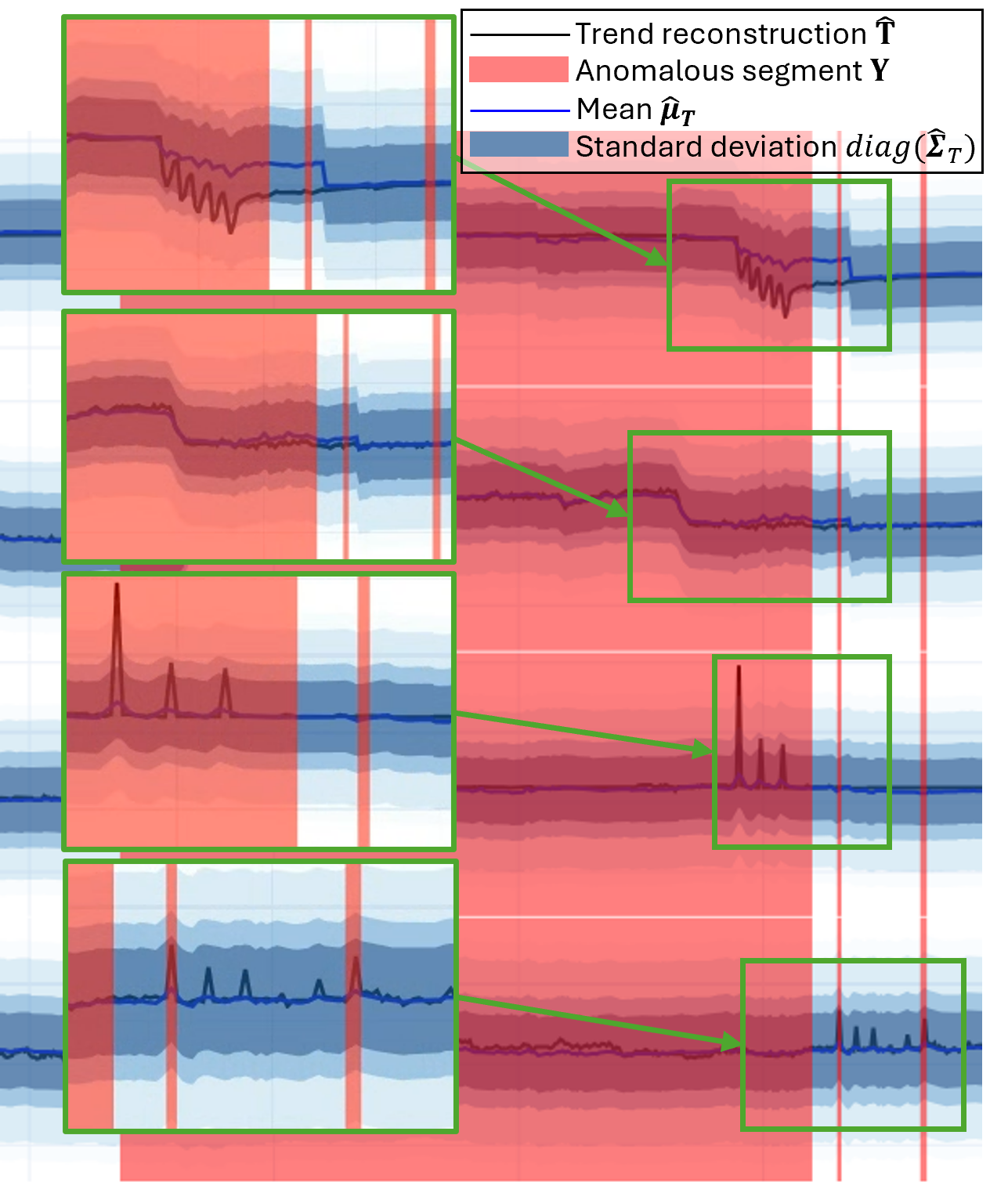}
        \caption{Exclude Detrending}
        \label{fig:exdetrend}
    \end{subfigure}
    \caption{Comparison of PSM features across model variants}
    \label{fig:stationarity}
\end{figure}

We conducted ablation studies by removing each module individually. \textbf{Exclude Embedding} removes the input standardization as described in Section \ref{secsub:se} and the denormalization steps in (\ref{denorm}) with \(\mathbf{E}_{\sigma}\) and \(\mathbf{E}_{\mu}\). \textbf{Exclude Detrending} removes the detrending procedure in (\ref{detrending}) and \(\mathbf{Z}_{TT}\) in (\ref{trendmu})-(\ref{trendcovariance}) is replaced by \(\mathbf{Z}^{(n_d)}_D\).  \textbf{Exclude FreqCNN} replaces $FreqCNN(\cdot)$ with a linear layer followed by a ReLU activation, projecting from $L$ to $d_m$ and back to $L$.
As shown in Table~\ref{tab:absf1}, each exclusion leads to a drop in TSAD performance, highlighting the benefits of all components. To illustrate the effect of detrending, the green boxes in Fig.~\ref{fig:stationarity} highlight the comparison. For the first two features, where trend-pattern anomalies are present, the estimated trend in Fig.~\ref{fig:ours_exdetrend} deviates more from the trend reconstruction than in the Exclude Detrending variant (Fig.~\ref{fig:ours_exdetrend}), resulting in clearer anomaly discrimination. For the remaining two features with abrupt spikes, the differences are minor. Overall, detrending improves anomaly clarity, especially for trend-related anomalies.

\subsubsection{Effect of FreqCNN Variants} The proposed FreqCNN is a crucial module that learns strong periodic features, making the model more effective in learning normal behavior that exhibits periodicity. In Table \ref{tab:freqcnn}, the symbol {\scriptsize\XSolidBrush}\text{ }denotes the change of specific sub-components within FreqCNN. To assess their impact, we have conducted ablation studies by changing the rFFT resolutions of \(\frac{L}{2^{l_f}}\) to a constant resolution of \(L\) and substituting the complex-valued CNN with a complex-valued MLP. In Table \ref{tab:freqcnn}, the use of \(\frac{L}{2^{l_f}}\) resolution rFFT and the CNN architecture improves TSAD performance, demonstrating the importance of FreqCNN’s design choices.

\begin{table*}[t]
\centering
\begin{minipage}{0.25\textwidth}
    \centering
        \caption{Model variants}
         
        \setlength{\tabcolsep}{2pt}
        \begin{tabular}{ccccc} \toprule
        \multicolumn{2}{c}{Model Variants}&MSL & SWAT & PSM\\ \midrule
        \multicolumn{2}{c}{Exclude Embedding}&0.218&0.838&\textbf{0.690}\\
        \multicolumn{2}{c}{Exclude Detrending}&0.262&0.843&0.649  \\ 
        \multicolumn{2}{c}{Exclude FreqCNN}&0.299&0.823&0.565 \\ 
        \multicolumn{2}{c}{\textbf{Ours}}&\textbf{0.310}& \textbf{0.850}&0.687\\
        \bottomrule
        \end{tabular}
                \label{tab:absf1}
\end{minipage}%
\hspace{0.045\textwidth} 
\begin{minipage}{0.3\textwidth}
    \centering
\caption{FreqCNN variants}
 
\setlength{\tabcolsep}{2pt}
\begin{tabular}{ccccc} \toprule
\(\frac{L}{2^{l_f}}\) resolution rFFT&CNN&MSL & SWAT & PSM \\
\midrule
{\scriptsize\XSolidBrush}&{\scriptsize\XSolidBrush}&0.238&0.831&0.662\\ 
{\scriptsize\XSolidBrush}&\checkmark&\textbf{0.312}&0.846&0.590 \\ 
\checkmark &{\scriptsize\XSolidBrush}&0.278&0.849&0.549\\
\checkmark & \checkmark&0.310& \textbf{0.850}&\textbf{0.687}\\
\bottomrule
\end{tabular}
\label{tab:freqcnn}
\end{minipage}
\hspace{0.025\textwidth} 
\begin{minipage}{0.35\textwidth}
    \centering
\caption{Anomaly criterion variants}
\setlength{\tabcolsep}{2pt}
 
\begin{tabular}{cccc} \toprule
Anomaly Criterion Variants&{MSL}&{SWAT}&{PSM}\\ \midrule
MSE&0.293&0.836&0.600\\ 
Exclude Summation&0.261&0.653&0.435\\
Exclude $\hat{\boldsymbol{\pi}}$&0.296&0.828&0.632\\
Exclude $\hat{\boldsymbol{\pi}}$ \& Summation&0.232&0.794&0.642\\
TrendLike (\textbf{Ours})&\textbf{0.310}& \textbf{0.850}&\textbf{0.687} \\
\bottomrule
\end{tabular}
\label{tab:abscriter}
\end{minipage}%
\end{table*}
\subsubsection{Effect of Different Anomaly Criterion} 
\begin{figure}[!h]
    \centering
    \includegraphics[width=0.85\linewidth]{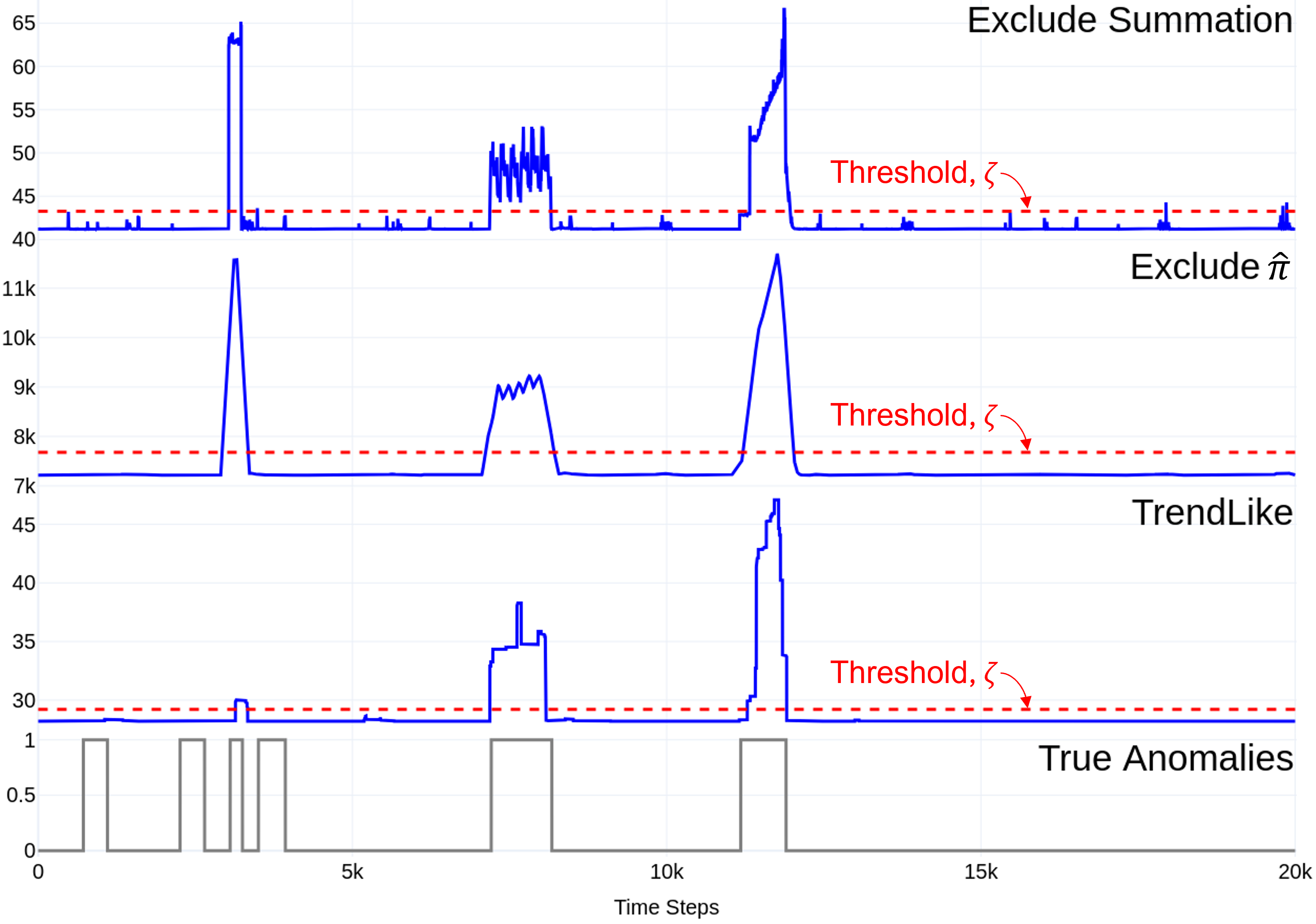} 
    \caption{Anomaly scores for a segment of the SWAT dataset}
    \label{scoringcriteria}
\end{figure}

\begin{figure}[!h]
    \centering
    \includegraphics[width=0.85\linewidth]{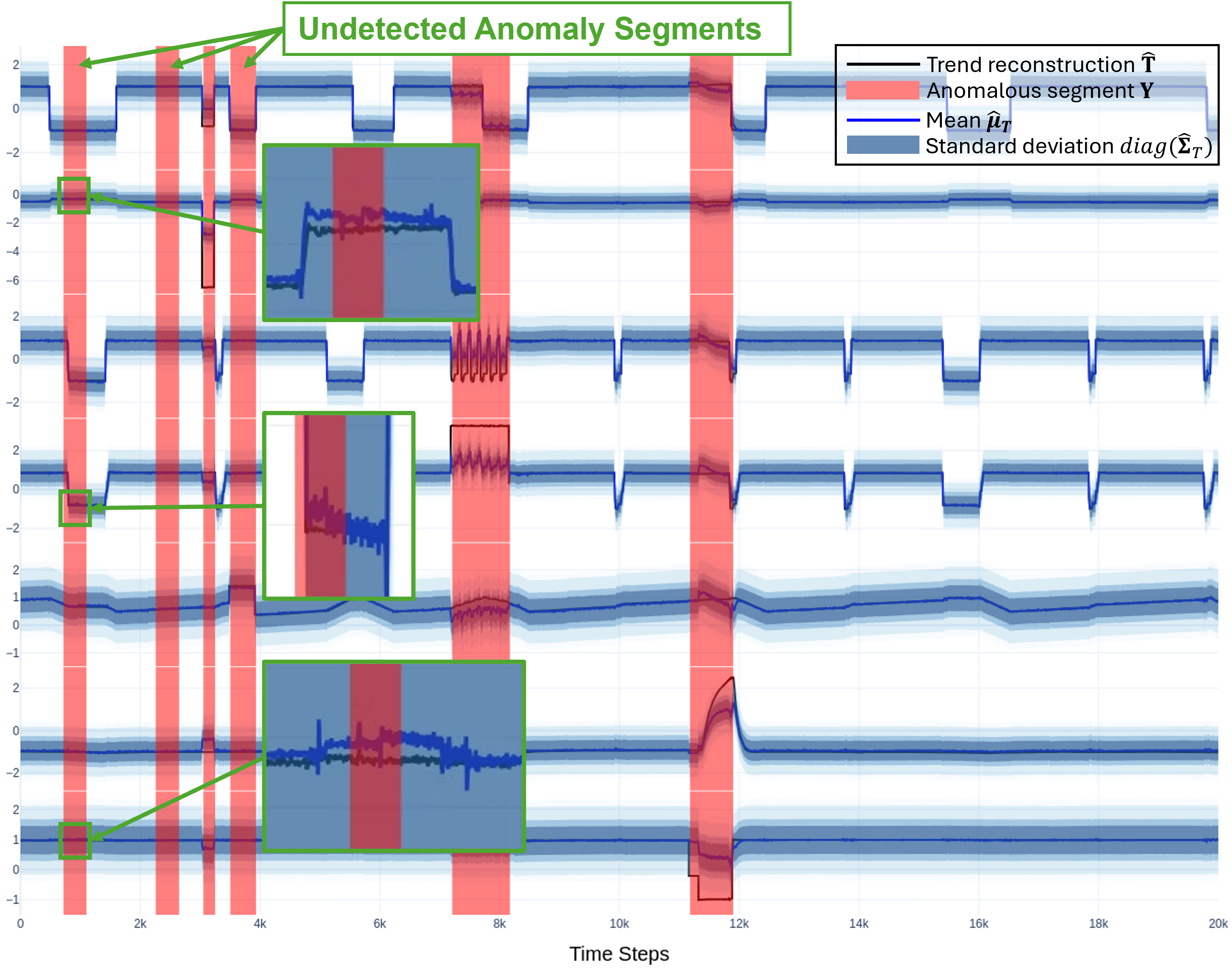} 
    \caption{Trend estimation for a segment of the SWAT dataset}
    \label{scoringcriteriaplots}
\end{figure}

This section analyzes the performance of the three anomaly criteria: \textbf{MSE} computes the anomaly score using the reconstruction loss in (\ref{lossrec}), where the ground-truth trend $\mathbf{T}$ is replaced with $\boldsymbol{\mu}_T$ during both training and anomaly score computation; \textbf{Exclude} $\hat{\boldsymbol{\pi}}$ removes the temporal vector from TrendLike in (\ref{trendlikeeq}) and from the likelihood loss in (\ref{likeloss}); \textbf{Exclude Summation} omits the summation over adjacent time steps in (\ref{trendlikeeq}). As illustrated in Fig.~\ref{scoringcriteria}, our proposed TrendLike method produces distinct and consistent anomaly scores. In contrast, variants that exclude the temporal vector tend to overestimate anomaly durations, yielding anomalous segments that exceed the ground-truth boundaries and thus increase false alarms, whereas those that exclude the summation exhibit random spikes even during normal segments. Table~\ref{tab:abscriter} shows that TrendLike consistently delivers superior results. While our anomaly criterion achieves the best overall performance, it still fails to detect certain anomalous segments, as illustrated in Fig.~\ref{scoringcriteria}, corresponding to the trend estimation in Fig.~\ref{scoringcriteriaplots}. The detected anomalies typically involve multiple features within the same time segment and are accompanied by abrupt changes, whereas undetected segments, highlighted in green, exhibit only minor deviations within the estimated trend distribution, indicating a limitation in capturing subtle anomalies.

\subsubsection{Threshold selection protocols}
For large-scale time series, labeled anomalies are often scarce due to high annotation costs and the rarity of anomalous events~\cite{xuanomaly,EvoAAE}, thus using labeled anomaly information for threshold selection~\cite{nspr,zong2018deep,xuanomaly} may be infeasible. Therefore,  we conduct an ablation study using three label-free methods: median thresholding ($\zeta = \text{median score}$), mean thresholding ($\zeta = \text{mean score}$), and a K-means-based approach ($\zeta = $ larger cluster centroid minus twice its standard deviation). Table~\ref{tab:thresholding} shows that thresholds computed solely from unlabeled data do not consistently achieve optimal performance. Nonetheless, the K-means approach remains competitive, with results on MSL, SWaT, and PSM close to the best F1 scores.

\begin{table}[H]
\setlength{\tabcolsep}{2pt}
        \centering
         
        \captionof{table}{Threshold values and F1-scores for different threshold selection protocols.}
        \begin{tabular}{c|cccccccccc} \toprule
        \multirow{2}{*}[-0.3em]{Methods}&\multicolumn{2}{c}{SMD}&\multicolumn{2}{c}{MSL}&\multicolumn{2}{c}{SMAP}&\multicolumn{2}{c}{SWAT}&\multicolumn{2}{c}{PSM}\\
        \cmidrule(lr){2-11}
        &$\zeta$&F1&$\zeta$&F1&$\zeta$&F1&$\zeta$&F1&$\zeta$&F1\\
        \midrule
        Median&21.04&0.988&30.44&0.246&14.05&0.293&28.18&0.217&13.85&0.438\\ 
        Mean&21.07&0.134&30.45&0.302&14.08&0.299&28.18&0.217&13.86&0.451\\
        K-means&21.00&0.080&30.46&0.300&13.95&0.252&29.39&0.840&13.97&0.660\\
        Best F1&22.10&0.203&30.46&0.310&14.50&0.378&29.20&0.850&14.00&0.687\\
        \bottomrule
        \end{tabular}\label{tab:thresholding}
\end{table}

\subsubsection{Hyperparameter Sensitivity Analysis}
\label{sec:hyper}
Each hyperparameter sensitivity analysis is conducted by varying one parameter over a predefined range while keeping all other parameters fixed, as listed in Table~\ref{tab:hyperparam_full}. Across datasets, the PSM dataset exhibits the greatest sensitivity to hyperparameters, potentially due to its higher anomaly ratio. Accordingly, hyperparameter tuning may improve model fitting and enhance anomaly detection performance in datasets with higher anomaly ratios. As shown in Fig. \ref{fig:hypersensitivity}, the sensitive hyperparameters are mainly \(L\), \(d_m\), and \(n_d\). For \(L\), shorter input lengths yield better performance, which declines as the input length increases. Meanwhile, increasing $d_m$ may improve performance, but there is a trade-off between accuracy and model complexity. Varying the number of $n_d$ layers allows the model to capture both local and global trends. For certain datasets, particularly SMAP and PSM, it has a significant impact, suggesting that proper tuning is necessary. The hyperparameter $g_{\text{cnn}}$ has little effect on most datasets except PSM, suggesting that larger values can be used to reduce computational complexity while maintaining accuracy. The remaining hyperparameter exhibits low sensitivity and can be set to constant values with minimal impact on performance. Overall, tuning the parameters $L$, $d_m$, and $n_d$ is generally sufficient to identify near-optimal hyperparameter settings for most datasets.

\newlength{\mysubfigwidth}
\setlength{\mysubfigwidth}{0.325\linewidth}
\begin{figure}[!h]
    \centering
        \begin{subfigure}[b]{0.325\linewidth}
        \centering
        \includegraphics[width=\mysubfigwidth]{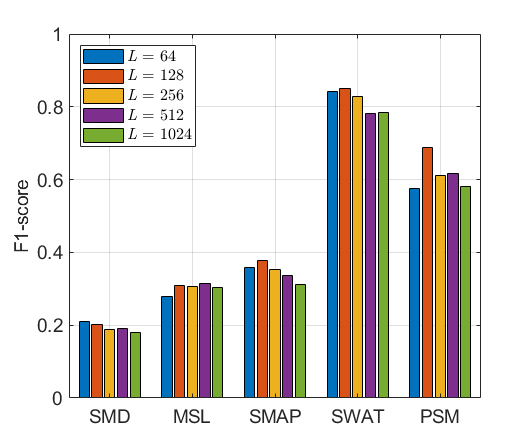}
        \caption{$L$}
    \end{subfigure} 
    \begin{subfigure}[b]{0.325\linewidth}
        \centering
        \includegraphics[width=\mysubfigwidth]{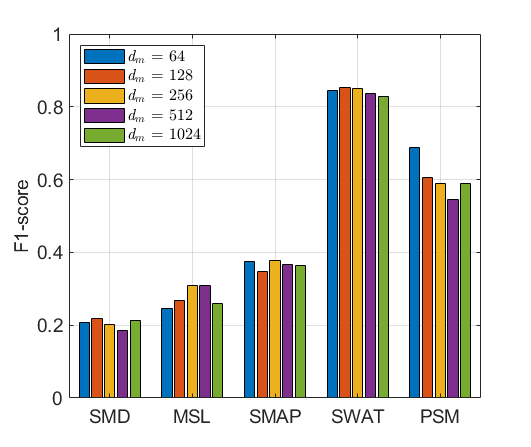}
        \caption{\(d_m\)}
        \label{fig:sub1}
    \end{subfigure}
    \begin{subfigure}[b]{0.325\linewidth}
        \centering
        \includegraphics[width=\mysubfigwidth]{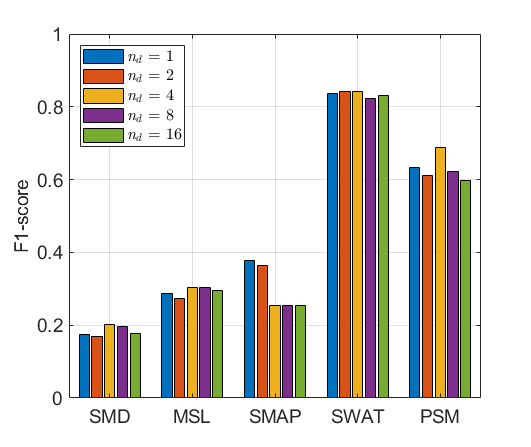}
        \caption{\(n_d\)}
        \label{fig:sub2}
    \end{subfigure}
    \begin{subfigure}[b]{0.325\linewidth}
        \centering
        \includegraphics[width=\mysubfigwidth]{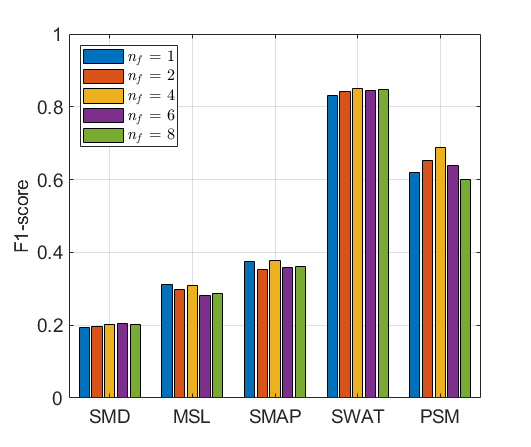}
        \caption{\(n_f\)}
        \label{fig:sub3}
    \end{subfigure}
    \begin{subfigure}[b]{0.325\linewidth}
        \centering
        \includegraphics[width=\mysubfigwidth]{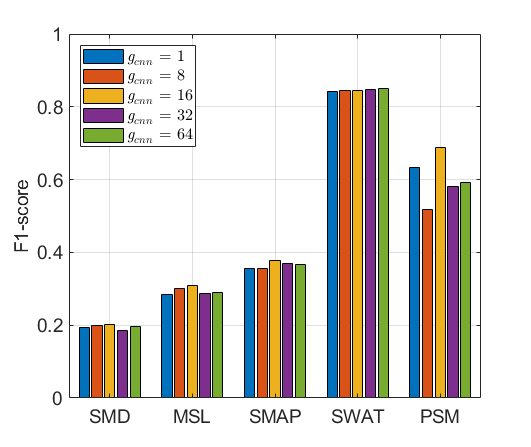}
        \caption{\(g_{cnn}\)}
        \label{gcnn}
    \end{subfigure}
    \begin{subfigure}[b]{0.325\linewidth}
        \centering
        \includegraphics[width=\mysubfigwidth]{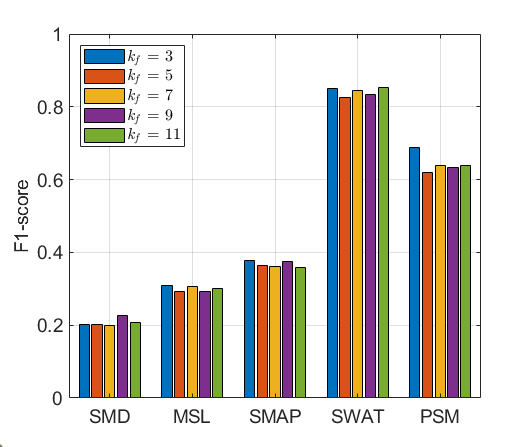}
        \caption{\(k_f\)}
        \label{fig:sub4}
    \end{subfigure}
    \begin{subfigure}[b]{0.325\linewidth}
        \centering
        \includegraphics[width=\mysubfigwidth]{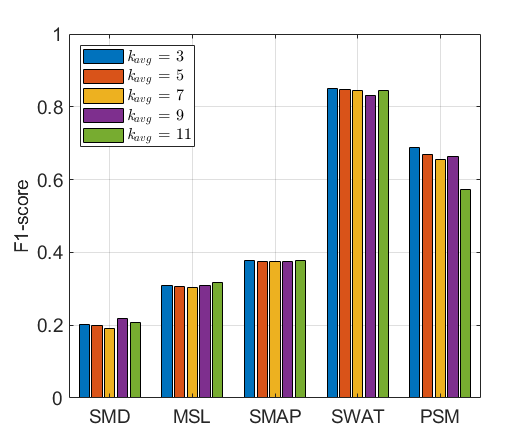}
        \caption{\(k_{avg}\)}
    \end{subfigure}
        \begin{subfigure}[b]{0.5\linewidth}
        \centering
        \includegraphics[width=1.15\mysubfigwidth]{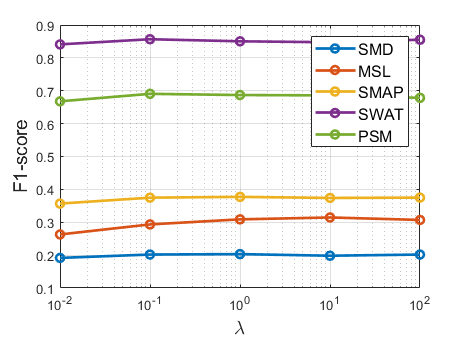}
        \caption{\(\lambda\)}
    \end{subfigure}  
    \caption{Hyperparameters sensitivity} \label{fig:hypersensitivity}
\end{figure}
\subsubsection{Complexity Analysis}
Table~\ref{tab:complexity} summarizes the training and inference computational complexity and runtime of the four baseline models (IF, LOF, TranAD, and NSPR) in comparison with our proposed model on the SWaT dataset. For consistency and fair comparison, all deep learning (DL) models, including TranAD, NSPR, and the proposed model, are trained with a fixed batch size of 64, whereas IF and LOF do not require batch-wise training. Runtime is reported as training time per epoch, and inference time is measured per sample, where each sample corresponds to a window of 100 time steps. Among the baseline methods, IF incurs the lowest training cost. Although LOF is more computationally expensive, it remains relatively efficient compared to deep learning (DL) approaches, which require multiple training epochs. For per-sample inference, IF achieves the fastest runtime due to its shallow tree traversal. In contrast, LOF is the slowest among the baselines, as it requires computing k-nearest neighbor distances and estimating local densities for each sample. DL models, including TranAD, NSPR, and our model, entail higher training costs but maintain competitive inference efficiency.

The TrendLike module is the primary computational bottleneck. In particular, the final covariance reconstruction $\mathbf{L}_{\sigma,l}\mathbf{L}_{\sigma,l}^\top$ introduces a computational complexity of $\mathcal{O}(L d_x^3)$. This term becomes the primary bottleneck when $d_x$ is large. TranAD~\cite{tranad} employs a standard Transformer architecture with attention complexity $\mathcal{O}(L^2 d_m)$, while NSPR adopts Performer~\cite{performer}, a linearized Transformer variant with attention complexity $\mathcal{O}(L r d_m)$, where $d_m \gg r$. All DL models include a feedforward network (FFN) with computational complexity $\mathcal{O}(L d_m^2)$. When $d_m \gg L, d_x$, the FFN term dominates across all methods, leading to similar overall computational costs. When $L \gg d_m, d_x$, TranAD scales quadratically with $L$ and becomes the most expensive, while NSPR remains the most efficient. Our method scales linearly with $L$ but complexity depending on feature dimension $d_x$. When $d_x \gg L, d_m$, our model becomes the most computationally expensive. In this regime, NSPR remains the most efficient, while the relative ordering between TranAD and our method depends on the specific values of $d_m$ and $d_x$. TranAD may incur higher practical runtime due to its two-stage training procedure, while its inference speed remains competitive. NSPR introduces overhead from stacking multiple linearized Transformer blocks, which increases runtime despite its lower theoretical complexity. In contrast, the TrendLike module is computed in a single forward pass, enabling efficient training and inference. Despite its additional computational cost, our approach remains competitive with existing methods.

\begin{table}[H]
\centering
\setlength{\tabcolsep}{3pt}
\caption{Models complexity on the SWAT dataset}
 
\begin{tabular}{c|c|c|c}
\toprule
&Training/&Inference/&Parameters \\

Models& runtime per epoch(s)&runtime per sample(ms)&(Million)\\
\midrule
IF\cite{liu2008isolation} & $\mathcal{O}(T \cdot \psi \cdot \log \psi )$/1.4& $\mathcal{O}(T \cdot \log \psi )$/0.05&0.011\\
LOF\cite{breunig2000lof} & $\mathcal{O}(M^2d_x)$/69.88& $\mathcal{O}(MLd_x)$/32.7&0.5\\
TranAD\cite{tranad} & $\mathcal{O}(L^2 d_\text{m})$/194.63& same as training/3.51&0.16\\
NSPR\cite{nspr} & $\mathcal{O}(L r d_m)$/34.47& same as training/17.9&0.39\\
\textbf{Ours} & $\mathcal{O}(L d_x^3)$/19.52& same as training/8.60&0.92\\
\bottomrule
\end{tabular}
\label{tab:complexity}
\vspace{1mm}
\footnotesize{
\parbox{\columnwidth}{%
$T$ = number of trees, $\psi$ = subsample size, $M$ = total length of the dataset%
}}
\end{table}
\subsubsection{Distribution variants} 
\begin{figure}[h]
    \centering
    \begin{subfigure}[b]{0.85\linewidth}
        \centering
        \includegraphics[width=\linewidth]{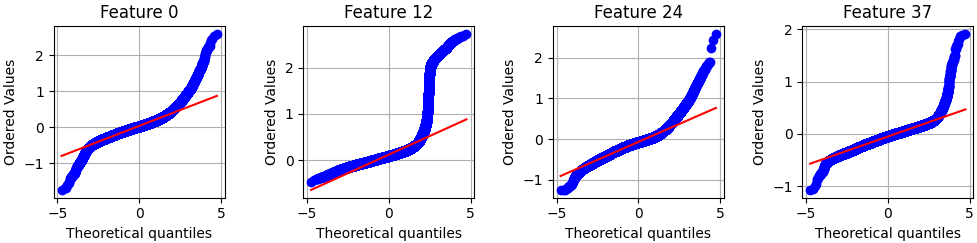}
        \caption{SMD}
        \label{fig:fullabnomral}
    \end{subfigure}
    \hfill
    \begin{subfigure}[b]{0.85\linewidth}
        \centering
\includegraphics[width=\linewidth]{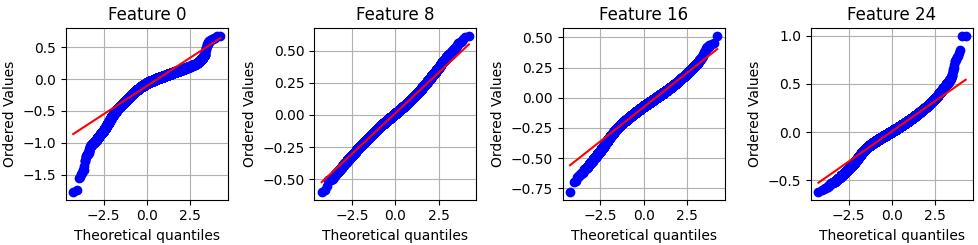}
        \caption{PSM}
        \label{fig:splitnormal}
    \end{subfigure}
    \caption{Q-Q plots of 4 features for SMD and PSM}
    \label{fig:qq_plot}
\end{figure}

To assess the impact of different distributional assumptions, we conduct an ablation study using four alternatives: an independent and identically distributed (i.i.d.) Gaussian distribution (G-dist), an i.i.d. Student’s t-distribution (t-dist), and quantile prediction. These variants are implemented by replacing the $\text{TrendEst}(\cdot)$ module without altering the overall model architecture. For i.i.d. G-dist, each element of the $L \times d_x$ time series is modeled independently by a Gaussian distribution, with the mean $\boldsymbol{\mu} \in \mathbb{R}^{L \times d_x}$ and standard deviation $\boldsymbol{\sigma} \in \mathbb{R}^{L \times d_x}$ estimated using 1DCNNs, the latter with a softplus activation. For the i.i.d. Student's $t$-distribution, an additional 1DCNN with a softplus activation is employed to estimate the degrees of freedom (DOF). Both variants are trained by maximizing the log-likelihood formulation. For quantile predictions, a 1DCNN estimates $q=100$ quantiles and is optimized using the pinball loss \cite{Wen2017AMQ}. The multivariate Student’s $t$-distribution (Multi t-dist) follows the same estimation procedure as the multivariate Gaussian distribution, with an additional 1DCNN using a softplus activation to estimate a time-dependent DOF parameter, producing an $L$-dimensional vector. For all variants, the computation of the temporal vector and detrended reconstruction losses remains unchanged. Table~\ref{tab:distributionabs} shows that the i.i.d. assumption consistently yields the poorest performance, as it fails to capture the feature dependencies inherent in real-world time series. In contrast, the multivariate and quantile-based distributions achieve comparable and substantially better results, with the multivariate formulation exhibiting greater robustness across datasets. These findings underscore the importance of distributional modeling in TSAD, suggesting the multivariate formulation as a reliable default choice, while the most suitable distribution may still depend on the characteristics of the dataset.
\begin{table}[H]
\setlength{\tabcolsep}{1.5pt}
        \centering
         
        \captionof{table}{Results for different distribution estimations}
        \begin{tabular}{c|cccccccccc} \toprule
        \multirow{2}{*}[-0.3em]{Distributions}&\multicolumn{2}{c}{SMD}&\multicolumn{2}{c}{MSL}&\multicolumn{2}{c}{SMAP}&\multicolumn{2}{c}{SWAT}&\multicolumn{2}{c}{PSM}\\
        \cmidrule(lr){2-11}
        &AUC&F1&AUC&F1&AUC&F1&AUC&F1&AUC&F1\\
        \midrule
        i.i.d G-dist&0.750&0.197&0.636&0.272&0.443&0.244&0.876&0.800&0.752&0.553\\\ 
        i.i.d t-dist&0.754&0.190&0.621&0.274&0.359&0.243&0.885&0.813&0.786&0.575\\ 
        Quantiles&\textbf{0.779}&\textbf{0.264}&0.646&0.296&0.423&0.242&\textbf{0.899}&0.796&0.733&0.564\\ 
        Multi t-dist&0.723&0.184&0.669&0.286&\textbf{0.727}&\textbf{0.398}&0.883&0.845&0.790&0.636\\ 
        \textbf{Ours}&0.722&0.203&\textbf{0.684}&\textbf{0.310}&0.658&0.377&0.884&\textbf{0.850}&\textbf{0.820}&\textbf{0.687}\\
        \bottomrule
        \end{tabular}\label{tab:distributionabs}
\end{table}
We further assess the validity of the Multi G-dist assumption on the SMD and PSM datasets, which are selected as representative cases of relatively poor and strong anomaly detection performance, respectively. Under a Multi G-dist, the standardized residuals $
\mathbf{r}_i=\hat{\boldsymbol{\Sigma}}_{T,i}^{-\frac{1}{2}}
\left(\hat{\mathbf{t}}_i-\hat{\boldsymbol{\mu}}_{T,i}\right),$ where \(i\) denotes the time step, are expected to follow a standard normal distribution. Therefore, we compute the standardized residuals, aggregate them across features at each time step, and pool the resulting values over all sliding windows of length 128 for evaluation using Q–Q plots. As shown in Fig.~\ref{fig:qq_plot}, the SMD dataset exhibits greater deviations from normality than the PSM dataset, indicating a poorer fit to the Multi G-dist. This is consistent with its comparatively poorer anomaly detection performance. Conversely, the PSM dataset exhibits good agreement with normality in the central region of the Q--Q plots. Overall, the Multi G-dist provides a reasonable approximation of the underlying trend distribution. Nevertheless, as the estimated trend distributions may deviate from the Multi G-dist assumption, feature-level interpretation is considered a secondary analysis and is applied only to segments identified as anomalous by the anomaly score.

\subsubsection{Anomaly Contamination Robustness}
Real-world datasets often contain undetected anomalies, which can distort the learned normal patterns \cite{couta}. To evaluate robustness under such conditions, we simulate anomaly contamination by injecting 10\%--50\% of abnormal samples from the original test set into the training data. All models are retrained on the contaminated training sets and evaluated under two settings. (i) Fig.~\ref{fig:fullabnomral} reports performance on the complete original test set, where the anomalous samples injected into the training set are also retained in the test set. This setting preserves the original test distribution and provides a comprehensive assessment of anomaly detection performance. (ii) Fig.~\ref{fig:splitnormal} reports performance on the remaining test samples after a subset of anomalous samples has been incorporated into the training set. This setting isolates the impact of training contamination while avoiding overlap between the training and test sets. Comparing these two settings mitigates bias from different train--test splits, as some anomalies are more difficult to detect. The detailed analysis is provided as follows:
\begin{itemize}
    \item For the SMD dataset, anomalies are subtle (Fig.~\ref{fig:limit}), and only 4.16\% of test time steps are anomalous. As a result, injecting additional anomalies has a limited impact on the training distribution, leading to similar performance across different contamination levels.

    \item For the MSL (Fig.~\ref{fig:mslest}) and SMAP (Fig.~\ref{fig:smapest}) datasets, anomaly contamination appears to have a limited impact on performance, as anomalies are typically erratic and manifest as spikes or abrupt changes. By explicitly modeling smooth long-term trends, our approach may better capture such deviations, contributing to improved robustness.

    \item For the SWAT dataset (Fig.~\ref{fig:swatest}), smooth and gradual anomalies may be partially treated as normal variations due to the model's emphasis on trend-level dynamics, increasing its susceptibility to contamination. Figs.~\ref{fig:swatnormal} and~\ref{fig:swatab} show that contamination causes over-generalization and reduced anomaly detectability.

    \item For the PSM dataset (Fig.~\ref{fig:psmest}), many anomalies fall within the estimated trend distribution. Consequently, training on contaminated data inflates the estimated variance to accommodate these fluctuations, as shown by the comparison between Figs.~\ref{fig:psmnormal} and~\ref{fig:psmab}.
\end{itemize}

Since evolving trends are a major source of time-series non-stationarity, we investigate whether sensitivity to imperfect trend estimation increases with dataset non-stationarity. The non-stationarity level is quantified using the average Wasserstein distance between consecutive Gaussian distributions parameterized by the sample mean and covariance of sliding windows with a length of 128 and a stride of 32, yielding scores of 0.198 (SMD), 0.187 (MSL), 0.266 (SMAP), 0.336 (SWAT), and 0.125 (PSM). To evaluate robustness under imperfect trend estimation, synthetic trend perturbations are introduced into the test set by injecting random level shifts and slope changes with varying magnitudes into randomly selected temporal segments, affecting 10\%--50\% of the time steps and features. As shown in Fig. 10c, although imperfect trend estimation causes performance degradation, the proposed model remains robust, with highly non-stationary datasets such as SWAT and SMAP exhibiting smaller performance degradation.

\begin{figure}[h]
    \centering
    \begin{subfigure}[b]{0.325\linewidth}
        \centering
        \includegraphics[width=\linewidth]{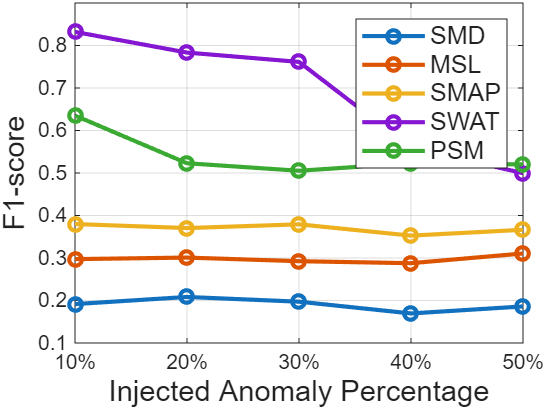}
        \caption{Full data}
        \label{fig:fullabnomral}
    \end{subfigure}
    \begin{subfigure}[b]{0.325\linewidth}
        \centering
        \includegraphics[width=\linewidth]{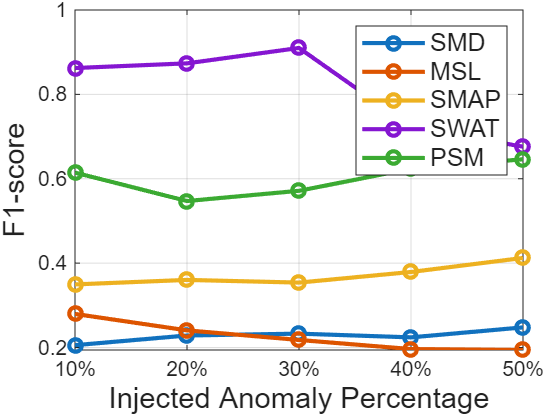}
        \caption{Partitioned data}
        \label{fig:splitnormal}
    \end{subfigure}
    \begin{subfigure}[b]{0.325\linewidth}
        \centering
        \includegraphics[width=\linewidth]{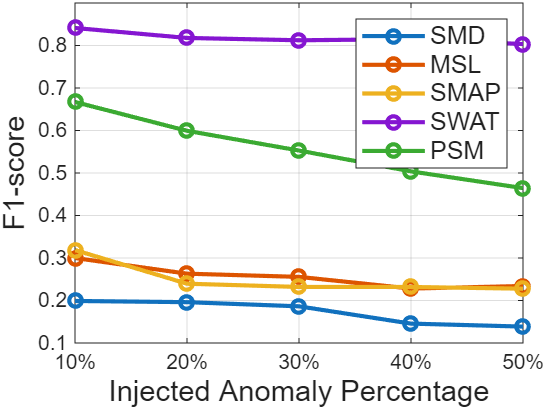}
        \caption{Synthetic anomaly}
        \label{fig:injectnoise}
    \end{subfigure}
    \caption{Performance under injected anomalies}
    \label{fig:noise}
\end{figure}
\begin{figure}[H]
    \centering
    \begin{subfigure}[b]{0.425\linewidth}
        \centering
        \includegraphics[width=\linewidth]{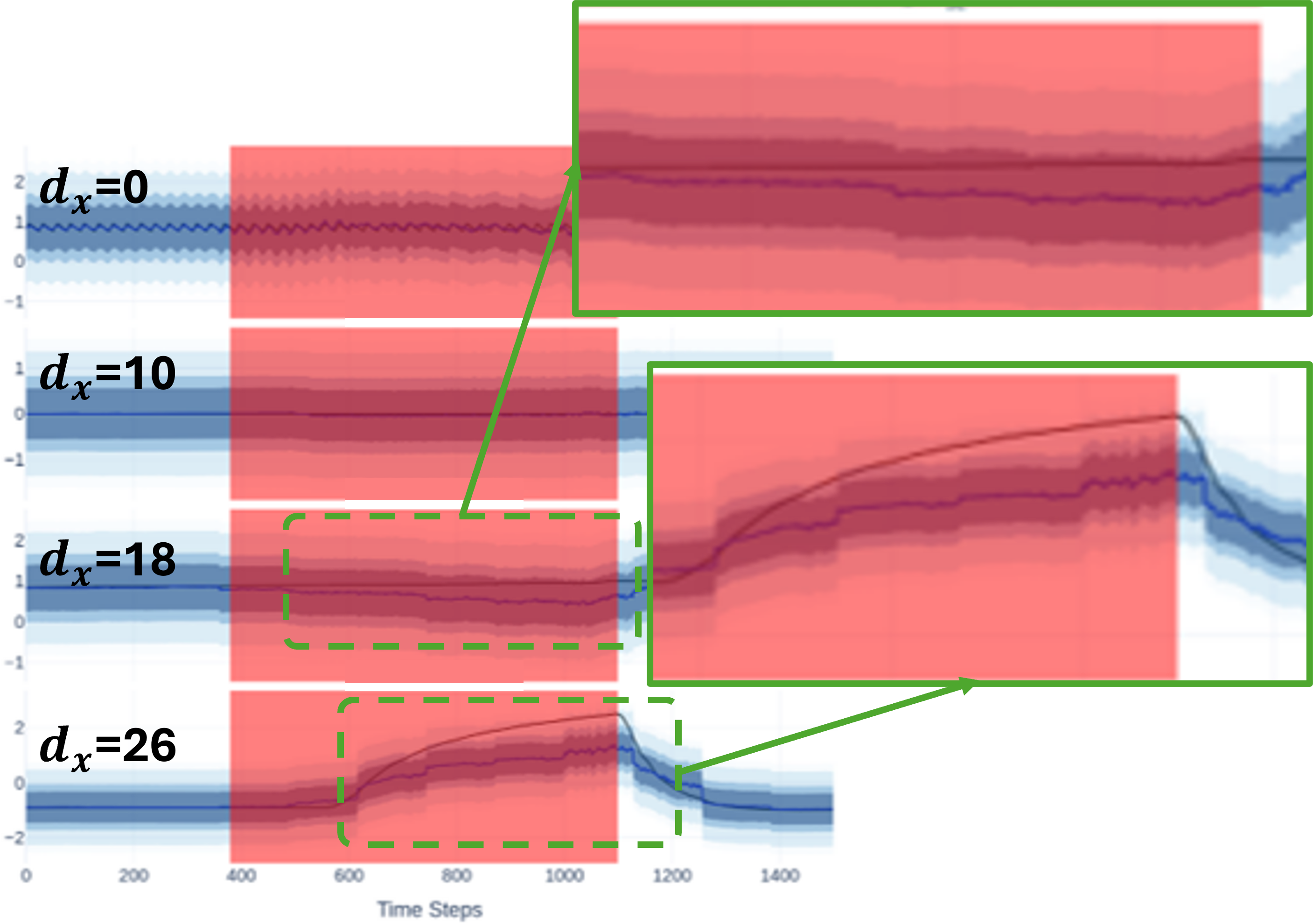}
        \caption{Anomaly-free (SWAT)}
        \label{fig:swatnormal}
    \end{subfigure}
    \hfill
    \begin{subfigure}[b]{0.425\linewidth}
        \centering
        \includegraphics[width=\linewidth]{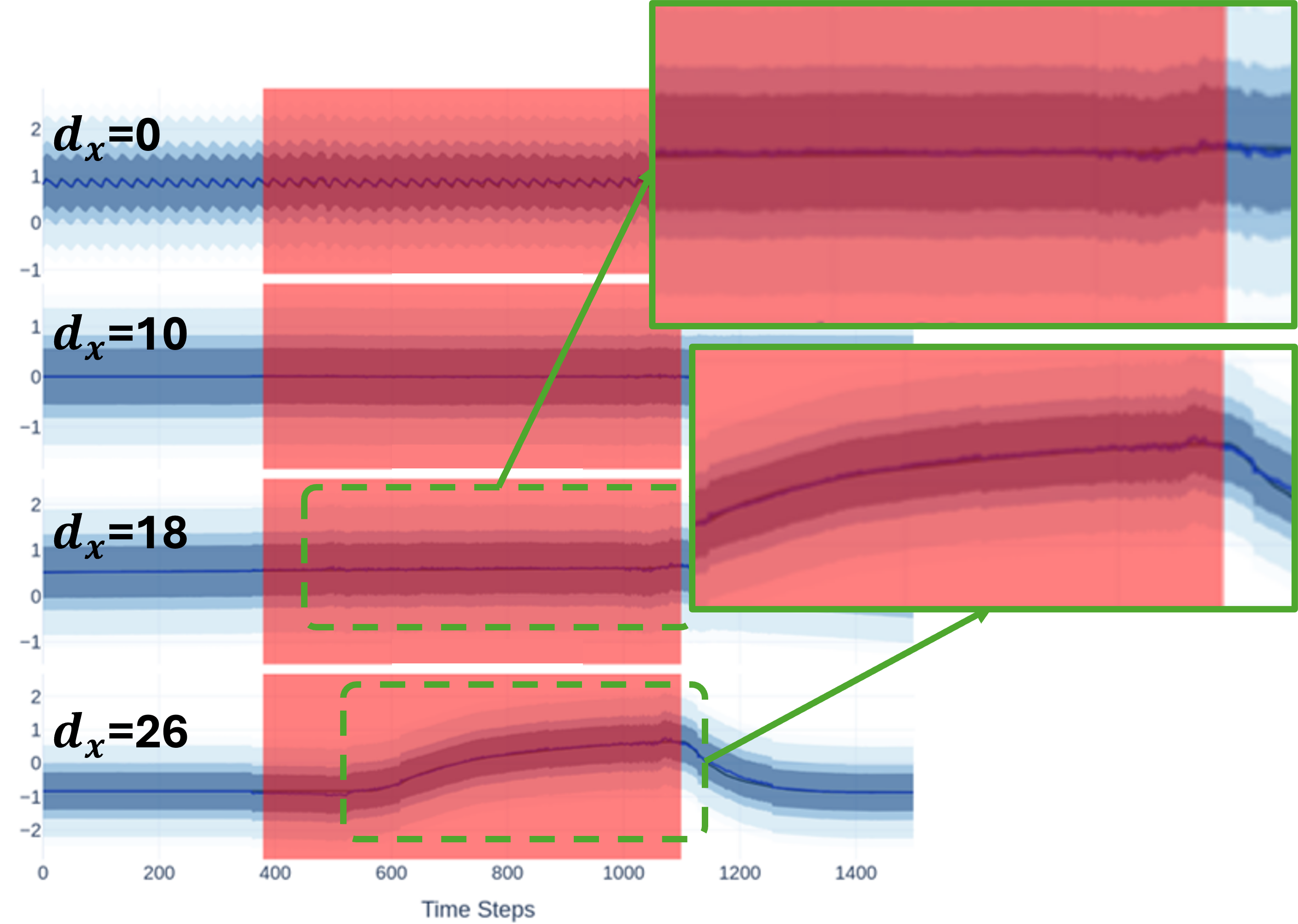}
        \caption{50\% Anomaly (SWAT)}
        \label{fig:swatab}
    \end{subfigure}
    \begin{subfigure}[b]{0.425\linewidth}
        \centering
        \includegraphics[width=\linewidth]{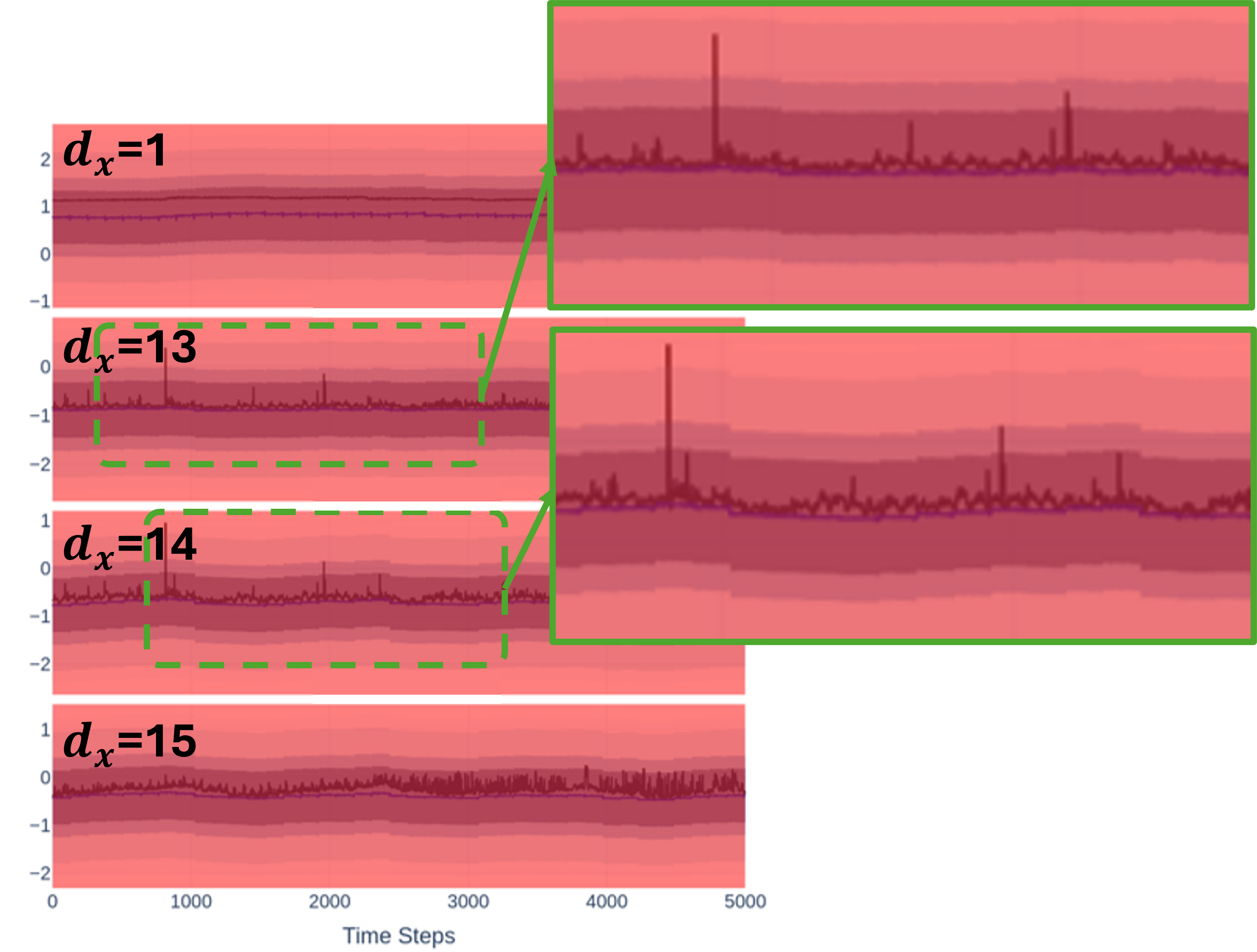}
        \caption{Anomaly-free (PSM)}
        \label{fig:psmnormal}
    \end{subfigure}
    \hfill
    \begin{subfigure}[b]{0.425\linewidth}
        \centering
        \includegraphics[width=\linewidth]{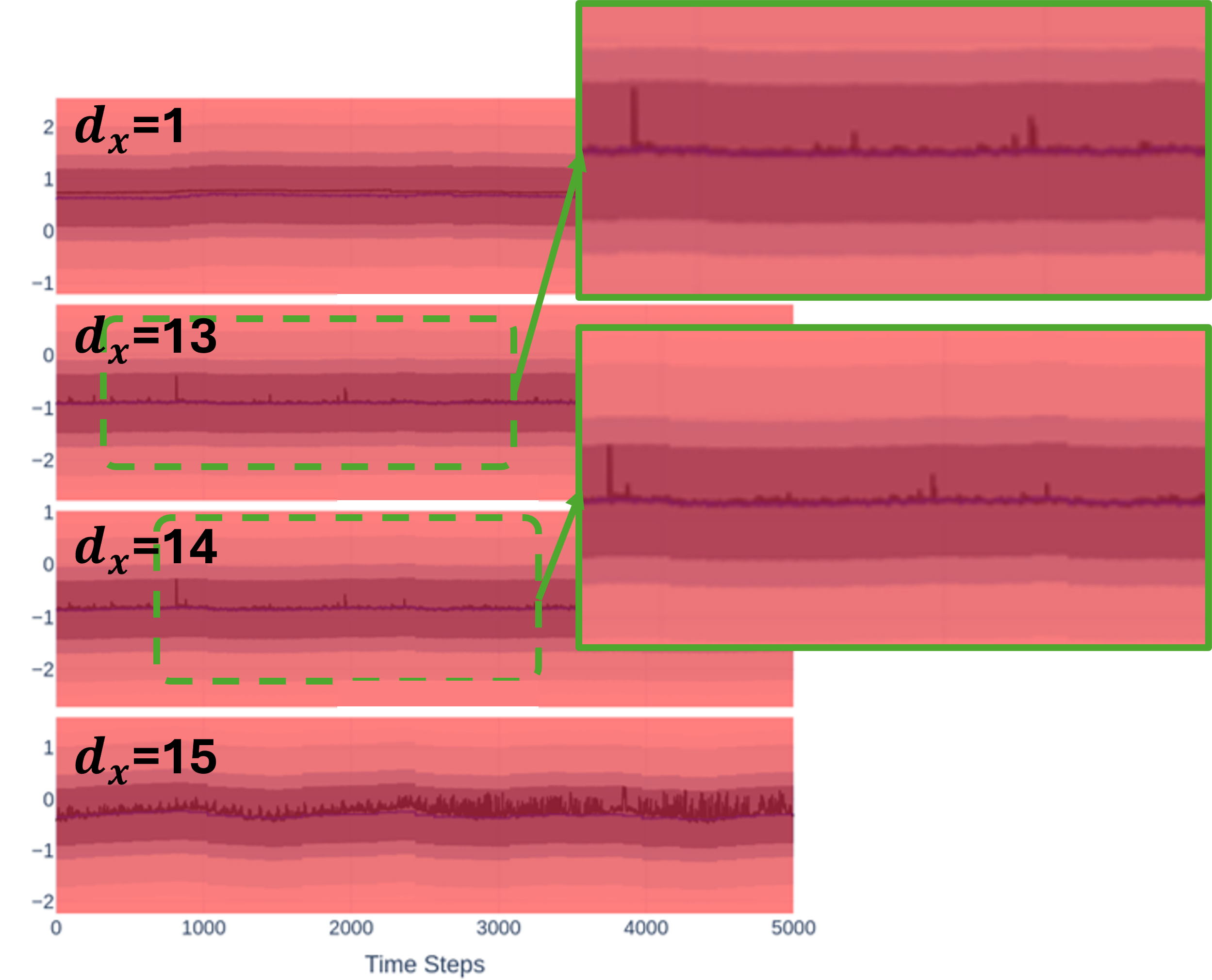}
        \caption{50\% Anomaly (PSM)}
        \label{fig:psmab}
    \end{subfigure}
    \caption{Trend estimation under anomaly contamination}
    \label{fig:noisecompare}
\end{figure}
\section{Conclusions}
\label{sec:conclude}
In conclusion, AdaTrendLike provides an effective framework for non-stationary time series by explicitly modeling trend and detrended components as distinct dominant elements. Hierarchical detrending facilitates estimation of their distribution and clearer separation from the ground truth. Complementing this, frequency-based convolutional neural networks further process the detrended component, improving anomaly detection performance. Together, these enable direct estimation of trend likelihood and provide feature-wise interpretability with actionable insights into which variables are affected by anomalous behavior. Despite these advances, datasets with subtle anomalies remain challenging, motivating future work to incorporate additional time-series components, such as cyclic and seasonal patterns, alternative distributional assumptions, and feature-wise anomaly scoring.

\bibliographystyle{IEEEtran}
\bibliography{myref} 
\end{document}